\documentclass{article} 
\usepackage[preprint]{colm2026_conference}

\usepackage{microtype}
\usepackage{hyperref}
\usepackage{url}
\usepackage{booktabs}
\usepackage{lineno}
\definecolor{darkblue}{rgb}{0, 0, 0.5}
\hypersetup{colorlinks=true, citecolor=darkblue, linkcolor=darkblue, urlcolor=darkblue}
\usepackage{tabularx}
\usepackage{relsize}

\usepackage{times}
\usepackage{latexsym}
\usepackage{amsmath}
\usepackage{amssymb}
\usepackage{array}      
\usepackage{geometry} 
\usepackage{lipsum}
\usepackage{caption}
\usepackage{amsmath}  

\usepackage[most]{tcolorbox}  
\usepackage{xcolor}           
\usepackage{verbatim}         
\usepackage{caption}          
\usepackage{graphicx}
\usepackage{subcaption}
\usepackage{adjustbox}    
\usepackage{amssymb}
\usepackage[utf8]{inputenc}
\usepackage{newunicodechar}

\usepackage{float}
\usepackage{stfloats}   
\usepackage{placeins}
\usepackage{enumitem}
\usepackage{subcaption}
\usepackage{wrapfig}

\newunicodechar{≈}{\approx}
\usepackage[T1]{fontenc}

\usepackage[utf8]{inputenc}
\usepackage{IEEEtrantools}

\usepackage{inconsolata}

\title{From Early Encoding to Late Suppression: Interpreting LLMs on Character Counting Tasks}

\author{%
  \textbf{Ayan Datta}$^{\dagger}$ \quad
  \textbf{Mounika Marreddy}$^{\ddagger}$ \quad
  \textbf{Alexander Mehler}$^{\ddagger}$ \quad
  \textbf{Zhixue Zhao}$^{\S}$ \quad
  \textbf{Radhika Mamidi}$^{\dagger}$ \\[0.5em]
  $^{\dagger}$IIIT Hyderabad \quad
  $^{\ddagger}$Goethe University, Frankfurt am Main, Germany \quad
  $^{\S}$University of Sheffield, UK \\[0.3em]
  \small\texttt{ayan.datta@research.iiit.ac.in} \quad
  \texttt{\{mmarredd, mehler\}@em.uni-frankfurt.de} \\
  \texttt{zhixue.zhao@sheffield.ac.uk} \quad
  \texttt{radhika.mamidi@iiit.ac.in}
}

\begin{document}

\ifcolmsubmission
\linenumbers
\fi

\maketitle

\begin{abstract}
Large language models (LLMs) continue to exhibit systematic failures on elementary symbolic tasks such as character counting in a word, depite excelling on complex benchmarks. Although this limitation has been noted in prior studies, the internal reasons remain underexplored. We use character counting (e.g., ``How many ‘p’s are in apple?”) as a minimal, fully controlled probe that isolates token-level reasoning from higher-level linguistic confounds. Using this setting, we uncover a striking and consistent phenomenon across modern architectures, including LLaMA, Qwen, and Gemma, and across both base and instruction-tuned variants: models frequently compute the correct answer internally yet fail to express it at the output layer.
Through mechanistic analysis combining probing classifiers, activation patching, logit lens analysis, and attention head tracing, we demonstrate that character-level information is already encoded in early and mid-layer representations. However, this information is not merely lost: it is actively attenuated by a small set of components in later layers, i.e., penultimate and final layer MLP and sometimes even the final layer Attention Heads. We identify these components as negative circuits: subnetworks that systematically downweight correct symbolic signals in favor of higher-probability but incorrect outputs. 
Our results lead to two central contributions. First, we show that symbolic reasoning failures in LLMs are not primarily due to missing representations or insufficient scale, but instead arise from structured interference within the model’s computation graph. This might explains why such errors persist and can even worsen under scaling and instruction tuning. Second, we provide the empirical evidence that LLM forward passes implement a form of competitive decoding, in which correct and incorrect hypotheses coexist and are dynamically reweighted, with final outputs determined by suppression as much as by amplification.
These findings carry important implications for interpretability and robustness: simple symbolic reasoning exposes systemic weaknesses in modern LLMs, underscoring the need for new design strategies that ensure information is not only encoded but also reliably used.

\end{abstract}

\section{Introduction}

\noindent
\begin{wrapfigure}{r}{0.45\textwidth}
    \vspace{-28pt}
    \centering
    \includegraphics[width=0.45\textwidth]{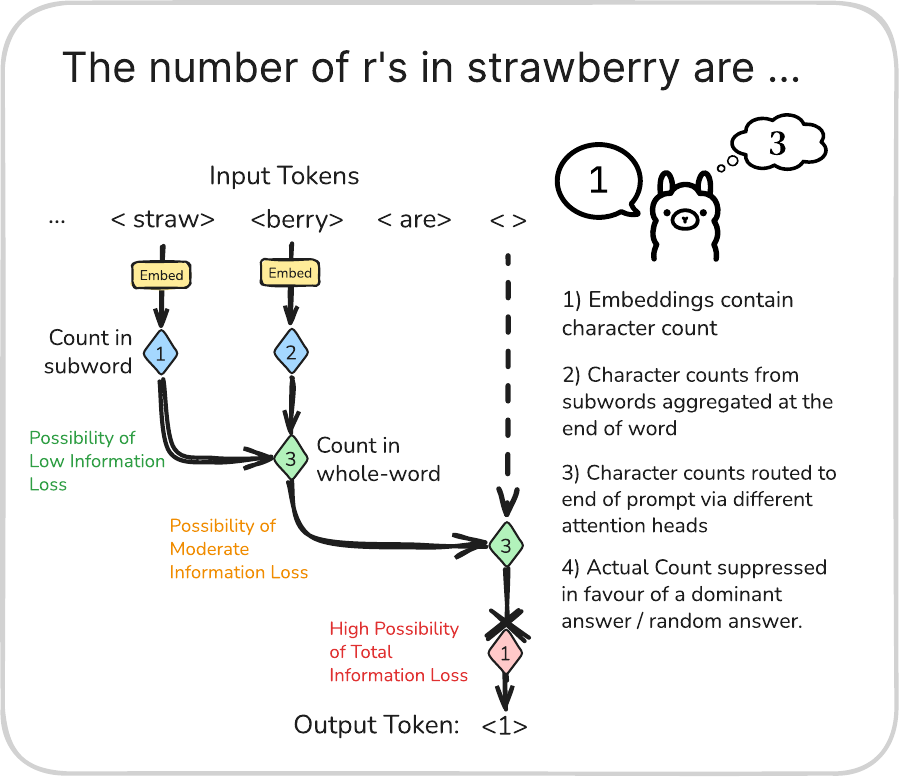}
    \caption{Overview of how character-level information flows through an LLM during the task.}
    \label{fig:teaser}
    \vspace{-20pt}
\end{wrapfigure}
Large language models (LLMs) \citep{achiam2023gpt, team2024gemini, chowdhery2022scaling, touvron2023llama, jiang2023mistral} achieve state-of-the-art performance on diverse natural language processing (NLP) tasks \citep{brown2020language,chung2024scaling}, yet they often fail on problems that appear trivial to humans, requiring only basic symbolic reasoning \citep{shin2024large}. Prior studies have documented such deficiencies in simple operations and reasoning, including letter containment and first/last character identification \citep{efrat2022lmentry}, spelling and character edits \citep{edman2024cute}, and character- and word-level manipulation tasks \citep{zhang2024large}. 
While these tasks are elementary, failures on them reveal systemic fragility: if models cannot reliably solve such atomic operations, their robustness on more complex reasoning tasks may rest on unstable foundations. 
Despite these observations, the mechanisms underlying such failures remain poorly understood. 

To this end, we revisit character counting as a controlled diagnostic task to probe this gap. Character counting is minimal yet stringent: solving it requires encoding fine-grained sub-token information and reliably propagating it through the model’s computation. Unlike surface-level accuracy, analyzing the internal components and dynamics provides insight into whether failures stem from representational absence or from interference during reasoning. We evaluate three recent model families, LLaMA 3.2 \citep{dubey2024llama}, Qwen 2.5 \citep{team2024qwen2}, and Gemma 2 \citep{gemmateam2024gemma2improvingopen}, in both base and instruction-tuned variants, using a dataset constructed from WordNet \citep{miller-1992-wordnet}. We employ probing classifiers, logit lens analysis, and activation patching to trace the dynamics of internal representation.
Our findings reveal structural patterns across families and scales:
\begin{itemize}
    \item LLMs are capable of capturing character level information in their residual stream.
    \item Attention heads do not specialize in routing character information.
    \item There exist components, particularly at the later layers, that suppress the probability of the correct answer. 
    \item Scaling the models from 2B to 9B parameters does not improve these issues, the same patterns remain even with larger models or with post training.
\end{itemize}

\section{Counting Task}
\label{sec:datagen}
To assess whether LLMs can process fine-grained character-level information, we design a controlled character-counting task. In this task, the model must determine how many times a specific target letter appears in a given word or phrase. To implement this, we construct natural language prompt templates where the correct count appears as the final token to be predicted. For example, the prompt ``The number of `i’s in `distribution' is ” elicits a probability distribution over plausible counts, enabling us to measure whether the model assigns high probability to the correct answer. To minimize biases from prompt phrasing or word selection, we employ multiple paraphrased templates, as shown in Table~\ref{tab:prompt_templates}. We design two sets of templates: one for base models (Table~\ref{tab:prompt_templates}) and another for instruction-tuned models (Appendix Table~\ref{fig:char_count_prompts}). Input words are sampled from WordNet \citep{miller-1992-wordnet} and scraped from the Wiktionary text dump \citep{wiktionary2024}. When generating prompt examples, we ensure that the distribution of target counts is balanced. By combining multiple templates with a diverse set of words, we reduce template- and word-specific biases, leading to a more reliable evaluation of the model’s character-counting ability.

We evaluate a range of base and instruction-tuned models on the same set of 10000 randomly generated dataset balanced across the final counts 1, 2, and 3 to establish their baseline performance on the task before analyzing the underlying reasons for failure. Since the prompts require only the prediction of the next token, we perform a single forward pass through the model and obtain probability distributions over the vocabulary. The token with the highest probability is then selected as the model’s predicted answer.

\begin{table}[!ht]
\centering
\scriptsize  
\begin{tabular}{|l|}
\hline
\textbf{Prompt Templates} \\ \hline
The number of \texttt{\textless target\_letter\textgreater}'s in \texttt{\textless count\_subject\textgreater} is \texttt{\textless count\textgreater} \\ \hline
Counting the letter \texttt{\textless target\_letter\textgreater} in \texttt{\textless count\_subject\textgreater} gives \texttt{\textless count\textgreater} \\ \hline
The number of \texttt{\textless target\_letter\textgreater}'s found in \texttt{\textless count\_subject\textgreater} is \texttt{\textless count\textgreater} \\ \hline
The \texttt{\textless target\_letter\textgreater} count for \texttt{\textless count\_subject\textgreater} equals \texttt{\textless count\textgreater} \\ \hline
The total number of \texttt{\textless target\_letter\textgreater}'s in \texttt{\textless count\_subject\textgreater} is \texttt{\textless count\textgreater} \\ \hline
\end{tabular}
\caption{Prompt templates for probing character count representations. For example: ``The number of p’s in apple is 2”, where \texttt{p} is the target letter and \texttt{apple} is the count\_subject.}
\label{tab:prompt_templates}
\end{table}

\section{Methodology}
This section details the experimental setup for evaluating LLMs on the character-level counting task. We systematically design controlled prompts and input variations to test the reliability of the model. Furthermore, we apply a series of interpretability techniques to identify the model components and computational pathways that are most responsible for generating the final counts.

\subsection{Probing Internal Representations}
\label{sec:probe_method}

We investigate whether these models encode character-count information within their internal representations, thereby testing the hypothesis that LLMs primarily operate at the token level and fail to capture sub-token features. If character-level information is present in hidden states, this would indicate that models retain fine-grained cues even if they do not reliably leverage them during generation. To distinguish between absence of encoding and failure of utilization, we conduct probing experiments to test whether the frequency of a given letter can be linearly decoded from internal representations.

\noindent\textbf{Probe Model.}
For each layer, we train a linear classifier without bias:

\begin{equation}
\label{eq:probe}
\mathbf{z} = W \mathbf{h}
\end{equation}

where $\mathbf{h} \in \mathbb{R}^{d}$ is the hidden representation, $W \in \mathbb{R}^{C \times d}$ is a learned weight matrix, and $C$ is the number of count classes. The probe is trained using cross-entropy loss,
\begin{equation}
\mathcal{L} = \mathrm{CrossEntropy}(\mathbf{z}, y),
\end{equation}
with predictions given by $\hat{y} = \arg\max_{c} z_c$.

\noindent\textbf{Dataset Construction.}
For each target letter, we construct a dataset from Wiktionary by grouping words according to the number of occurrences of the letter. Starting from count 1, we sample 1{,}000 words per count and continue until fewer than 1{,}000 examples are available. We additionally include 1{,}000 randomly sampled words with zero occurrences. For each word and letter, we generate inputs using all templates and extract hidden states across all layers. When a word is split into multiple tokens, we use the hidden state of the final token to obtain a single representation unless otherwise specified.

\noindent\textbf{Probing Settings.}
We consider three settings:
\begin{itemize}[leftmargin=*, itemsep=0.2em]
    \item \textbf{Token-level (individual tokens:} Predict the count of the target letter within individual tokens (subword units). Splits are constructed such that token types are disjoint across train, validation, and test sets.
    \item \textbf{Word-level (final word token):} Predict the total count of the target letter in the word using the final token representation. Word types are disjoint across splits.
    \item \textbf{Prompt-level (final prompt token):} Predict the total count using the final token of the full prompt. Word types remain disjoint across splits.
\end{itemize}

\noindent\textbf{Training Procedure.}
We use a 70\%-15\%-15\% train-dev-test split with partitioning over types rather than instances: no token (token-level) or word (word- and prompt-level) appears in more than one split. We generate prompts using all of the templates. Probes are trained with Adam \citep{kingma2017adammethodstochasticoptimization}, using an initial learning rate of $10^{-3}$, which is annealed by a factor of 0.5 when validation accuracy does not improve. Early stopping is based on validation performance, and results are reported on the test set.

We do not employ control tasks \citep{hewitt-liang-2019-designing} as our 
dataset construction already structurally prevents memorization; we discuss 
this in detail in Appendix~\ref{app:control_tasks}.

\subsection{Localizing Components}
We attempt to localize task behavior to different components of the model in order to identify the source of poor counting performance.

\subsubsection{Performance Evaluation}
To attribute changes in model behavior to specific internal components, we require evaluation metrics beyond accuracy. Accuracy alone is discrete and does not capture the model’s confidence or relative preference among competing tokens. We therefore adopt a logit-difference metric ($\Delta$) from \citep{wang2023interpretability, NEURIPS2023_34e1dbe9}, which measures how strongly the model favors the correct token relative to plausible alternatives as a linear function of the residual stream:
\begin{equation}
    \Delta = L(\text{correct token}) - \mathbb{E}_{t \sim T}[L(t)]
\end{equation}
\begin{equation}
    L(t) = (W_U \cdot \mathbf{h})[\text{VocabularyIndex}(t)]
\end{equation}
Here, $L(t)$ denotes the logit of token $t$, obtained by projecting the 
final hidden state $\mathbf{h}$ with the unembedding matrix $W_U$. $T$ is 
the set of the top-$N$ highest-probability incorrect tokens (we use $N=2$). 
Since the answer space dominantly consists of the tokens $\{1, 2, 3\}$, this in 
practice reduces to the logit of the correct count minus the mean logit 
over the remaining two candidates.

The $\Delta$ metric provides a continuous, interpretable measure of preference 
for the correct answer. Values of $\Delta > 0$ indicate that the correct token 
is favored over alternatives, and larger values correspond to stronger 
preference. Values of $\Delta < 0$ indicate that the correct token is less 
likely than the competing alternatives, reflecting a preference against the 
correct answer. Crucially, $\Delta$ is linearly related to the residual stream, 
making it well-suited for attributing behavior to internal components.

In addition, we introduce a probability-gap metric to measure how the correct answer is suppressed relative to the model’s observed prediction:
\begin{equation}
    \Delta_p = P(\text{observed}) - P(\text{correct})
\end{equation}
where probabilities are obtained by applying a softmax over logits. This metric captures the extent to which probability mass is shifted away from the correct token toward an incorrect alternative.

Unlike $\Delta$, the probability-gap metric is not linearly related to the residual stream due to the softmax transformation. However, it provides a 
normalized and more interpretable scale, allowing comparisons in terms of probabilities. The two metrics serve complementary roles. We use $\Delta$ 
to track whether the model is building preference for the correct answer across layers and components, regardless of what happens to other token logits. This is important because we want to isolate the signal for the correct answer without it being confounded by global logit shifts. We then 
use $\Delta_p$ to quantify suppression: by contextualizing the correct token's probability against the observed prediction, it captures how much probability mass is actively redirected away from the correct answer toward competing tokens. 
Together, the two metrics provide complementary 
information \citep{wiegreffe2025answerassembleaceunderstanding}, with $\Delta$ tracking the buildup of evidence for the correct answer and $\Delta_p$ revealing where and by how much that evidence is overridden.

We note that $\Delta$ is closely related to cross-entropy loss: taking 
differences in cross-entropy between two tokens reduces to their logit 
difference, as the softmax normalizer cancels (see Appendix~\ref{app:cross_entropy}).

\subsubsection{Logit Lens}
\label{sec:logit_lens_method}
To trace how predictions evolve through the network, we apply the Logit Lens \citep{nostalgebraist2020logitlens}, projecting intermediate hidden states onto the vocabulary space using the model’s unembedding matrix. At a given layer index $i$, we project hidden states from two different locations, $i_\text{pre}$ (residual stream right before the attention layer) and $i_\text{mid}$ (residual stream right before the MLP Layer). We also use the final layer's hidden state ($\text{final}_{post}$). Given an hidden state $\textbf{h}$ at layer $i$ either from the $\text{pre}$ position or the $\text{post}$ position, we first ensure the vector is normalise. This means ensuring it is the output of the model's corresponding LayerNorm \citep{ba2016layernormalization} or RMSNorm \citep{zhang2019rootmeansquarelayer} layer. If $\textbf{h}$ was not the output of a normalization layer, we apply the corresponding normalization to obtain $\textbf{h}_\text{norm}$. We compute the corresponding Logit values as such:
\begin{equation}
    L_{\text{intermediate}}(t) = (W_U \cdot \mathbf{h}_{\text{norm}})[\text{VI}(t)]
    \label{eq:logit_intermediate}
\end{equation}

We then use the logit values from $L_\text{intermediate}$ to compute $\Delta_p$ across different layer positions and VI is the vocabulary Index. This evolution of $\Delta_p$ values helps track at what point in the model, does the correct answer get suppressed.

\subsubsection{Activation patching} 
\label{sec:act-patch-method}

\begin{table*}[hbt]
\centering
\begin{tabular}{|l|p{5cm}|p{5cm}|}
\hline
\textbf{Corruption Type} & \textbf{Clean Prompt} & \textbf{Corrupted Prompt} \\ \hline
Word Corruption & The number of a's in \textbf{apple} is & The number of a's in \textbf{ada} is \\ \hline
Letter Corruption & The number of \textbf{a}'s in apple is & The number of \textbf{p}'s in apple is \\ \hline
Word+Letter Corruption & The number of \textbf{a}'s in \textbf{apple} is & The number of \textbf{p}'s in \textbf{ada} is \\ \hline
\end{tabular}
\caption{Examples of clean and corrupted prompts used in activation patching. Each corruption alters the expected letter count while maintaining the prompt structure.}
\label{tab:corruptions-table2}
\end{table*}

Activation patching, also referred to in the literature as causal tracing \citep{meng2022locating}, causal mediation analysis \citep{vig2020investigating}, or interchange intervention \citep{geiger2020neural} is a widely used interpretability technique for localizing causal mechanisms in neural networks. The method involves systematically overwriting hidden activations from a ``corrupted” input with cached activations from a ``clean” input and observing changes in the model’s predictions. By analyzing the degree to which the clean behavior is restored, one can identify which layers, neurons, or subspaces are causally responsible for the output \citep{zhang2023towards, heimersheim2024use}. Variants of this approach have been extensively applied across model interpretability studies \citep{geiger2021causal, soulos2019discovering, hase2023does, finlayson2021causal, wang2022interpretability, chan2022causal, hanna2023does,conmy2023towards}.

Building on this framework, we design our experiments using a controlled set of prompts. We use our data generator from Section \ref{sec:datagen} to create clean prompts and corrupted versions by altering either the word, target letter, or both, as shown in Table~\ref{tab:corruptions-table2}. For each clean-corrupted prompt pair, we run the clean prompt through the model, save the intermediate hidden states, and measure performance $\Delta_\text{clean}$. We then do the same for the corrupted prompt to obtain the corrupted hidden states, but use the clean prompt’s correct answer to measure $\Delta_\text{corrupted}$. To ensure meaningful comparisons, we only select prompt pairs where $\Delta_\text{clean} - \Delta_\text{corrupted} > 0.5$ and $\Delta_\text{clean} > 0$, which provides a proper range for evaluating performance restoration and ranking components according to their contribution to correct or incorrect outputs.

With these prompt pairs defined, we apply the activation patching procedure. For every hidden state at layer $L$ and token position $T$ in the clean prompt’s activations, we run the corrupted prompt up to the corresponding hidden state and replace (``patch in'') the clean hidden state. We then measure the new performance $\Delta_\text{patched}$ and quantify the restoration using the normalized metric:

\begin{equation}
P_{\text{restored}} =
\frac{
\Delta_{\text{patched}} - \Delta_{\text{corrupted}}
}{
\Delta_{\text{clean}} - \Delta_{\text{corrupted}}
}
\end{equation}

We use $\Delta$ rather than $\Delta_p$ for the activation patching metric for two reasons. First, as noted above, $\Delta$ is linearly related to the residual stream, which makes it directly interpretable as a measure of each component's additive contribution to the model's preference for the correct token. Second, probabilities are poorly suited for measuring restoration: the softmax transformation compresses probability mass non-linearly, causing probabilities to decay very rapidly once a token falls out of the top predictions. This means that $\Delta_p$ is disproportionately sensitive to small logit changes in regions where the correct token is already suppressed, producing a noisy and poorly calibrated restoration signal. The logit scale, being unbounded and logarithmic relative to probability, varies more smoothly across the range of patching conditions and yields a more stable normalized restoration score. We therefore reserve $\Delta_p$ for quantifying suppression at the output layer, where its normalized scale aids interpretability, and rely on $\Delta$ throughout the activation patching analysis.

This metric yields $P_\text{restored}$ for each layer and token position. A value of $P_\text{restored} = 0$ indicates that the component does not contribute to the task; $P_\text{restored} > 0$ indicates that it increases the probability of the correct answer; and $P_\text{restored} < 0$ indicates that it suppresses the probability, thereby hindering the correct answer.

To obtain comprehensive insights, we perform patching at multiple levels: (i) residual outputs of each layer, (ii) individual attention heads, and (iii) MLP sublayers, across all token positions. This multi-level intervention allows us to precisely localize causal contributions.

Finally, to aggregate results across different templates and token lengths, we compute performance restoration scores for each component at different token positions and average them. To prevent small or near-zero contributions from diluting strong effects, we adopt a weighted mean where each score is weighted by its absolute value:
Given a set of scores $S = \{s_1, s_2, ... s_n\}$
\begin{equation}\text{Aggregate}(S) = \mathlarger{\frac{\sum_{i=0}^{n} |s_i| \cdot s_i}{\sum_{i=0}^{n} |s_i|}}\end{equation}


For word corruptions, we aggregate across positions before, at, and after the target word. For letter corruptions, we aggregate before, at, and after the target letter. For combined word+letter corruptions, we further distinguish positions between the letter and word tokens. We ensure that all templates satisfy $\text{letter start} < \text{word start}$, enabling consistent alignment across conditions.

\subsection{Models and Implementation Details}
We perform our experiments over a range of recent open-weight large language models, covering both base (pretrained) and instruction-tuned variants. From the Qwen-2.5 family~\citep{team2024qwen2}, we include the 3B and 7B parameter models (Qwen-2.5-3B, Qwen-2.5-7B) along with their instruction-tuned counterparts (Qwen-2.5-3B-Instruct, Qwen-2.5-7B-Instruct). From Gemma-2~\citep{gemmateam2024gemma2improvingopen}, we consider the 2B and 9B models (gemma-2-2b, gemma-2-9b) and their aligned instruction-tuned versions (gemma-2-2b-it, gemma-2-9b-it). For LLaMA-3.2~\citep{dubey2024llama}, we focus on the 3B parameter model, evaluating both the base (LLaMA-3.2-3B) and instruction-tuned (LLaMA-3.2-3B-Instruct) variants. We also include an equivalent bigger model, LLaMA-3.1-8B and the instruction-tuned LLaMA-3.1-8B-Instruct \citep{grattafiori2024llama3herdmodels}.
All checkpoints are obtained from Hugging Face’s model hub~\citep{wolf2020transformers}. All experiments are implemented using the TransformerLens interpretability library. The models are executed on a computing setup equipped with an NVIDIA RTX 6000 Ada GPU, with 48 GB of VRAM.

\section{Results}

\subsection{Overall Performance}
~\label{sec:overall_perf}
\begin{wrapfigure}{r}{0.42\textwidth}
    \centering
    \vspace{-10pt}\includegraphics[width=0.40\textwidth,trim={0 2mm 0 2mm},clip]{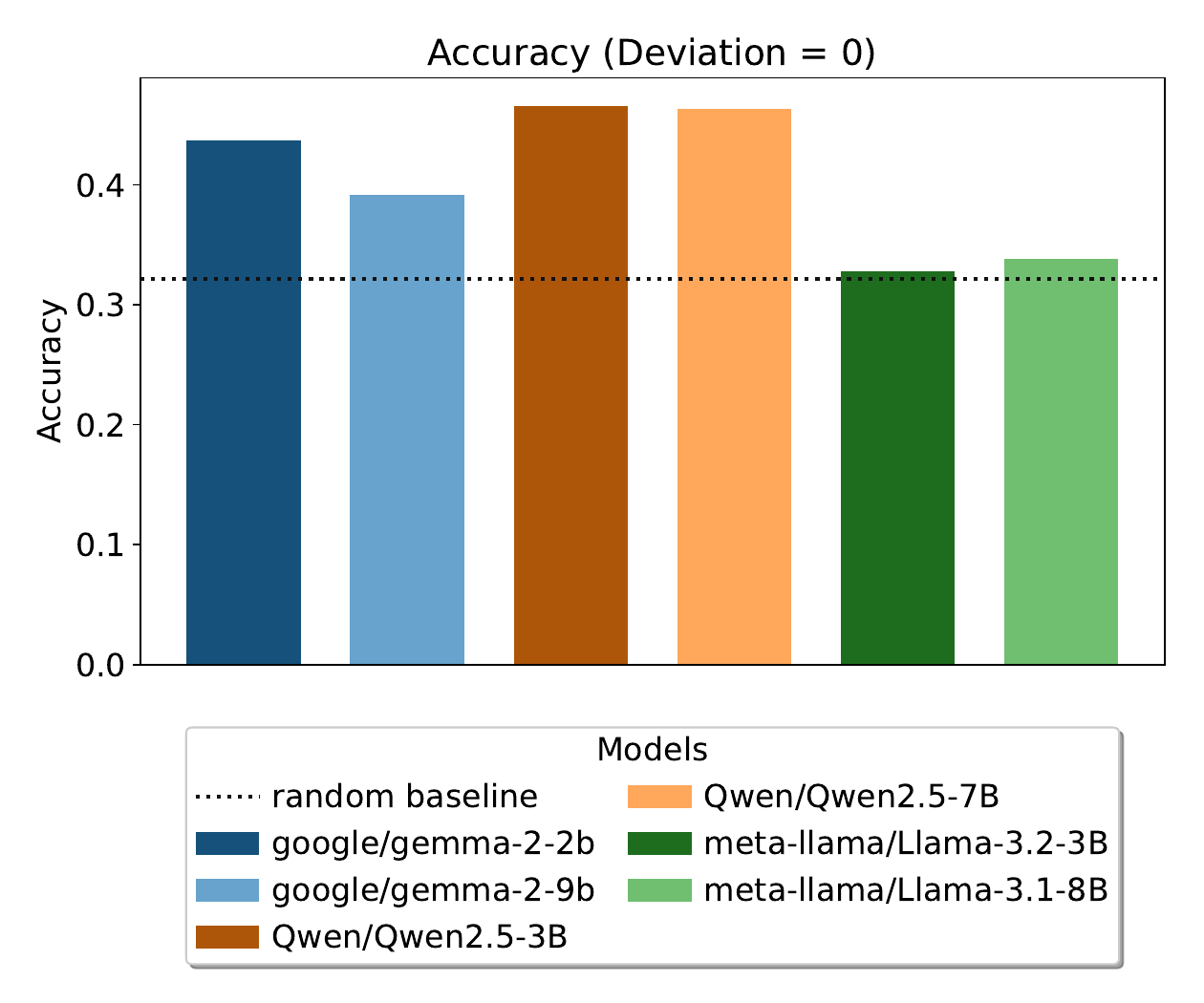}
    \caption{Accuracy of predicted letter counts. The dotted line indicates the random baseline accuracy. We calculate the proportion of samples where deviation is zero}
    \label{fig:perf_plot}
    \vspace{-20pt}
\end{wrapfigure}

\noindent\textbf{LLMs are unreliable at character counting.}
Figure~\ref{fig:perf_plot} reports the overall accuracy over all the models studied along with a random prediction baseline. We observe substantial error rates, with a large proportion of predictions being incorrect. Notably, even relatively recent architectures such as LLaMA-3.2, Qwen-2.5, and Gemma-2 fail to achieve reliable performance \textbf{achieving close to the random baseline's 33\% Accuracy}.

 These results confirm and extend prior findings on the limitations of LLMs in character-level tasks, suggesting that precise letter counting remains a systematic weakness rather than an isolated failure. To better understand these failures, we inspect confusion matrices (Appendix~\ref{app:confusion_matrices}, where we present results for LLaMA-3.2-3B and Qwen2.5-3B as representative examples of the two dominant failure modes) and observe that models adopt simple, degenerate answer strategies rather than performing true counting. \textbf{Gemma and Qwen variants overwhelmingly predict a fixed count (typically ``1''), while
LLaMA models exhibit near-uniform predictions across counts.} These behaviors explain both the poor overall performance and the patterns observed in Figure~\ref{fig:perf_plot}. In particular, the near-random strategy employed by LLaMA accounts for its accuracy being close to the random baseline. We discuss both strategies and their implications in detail in Appendix~\ref{app:confusion_matrices}.

We provide a detailed performance breakdown across factors such as letter type (vowel vs.\ consonant) and positional effects in Appendix~\ref{app:perf_analysis}.

\subsection{Probing Tasks}
\label{sec:probing}
\noindent
\begin{wrapfigure}{r}{0.40\textwidth}
    \centering
    \vspace{-6pt}

    \begin{subfigure}{\linewidth}
        \centering
        \includegraphics[width=\linewidth]{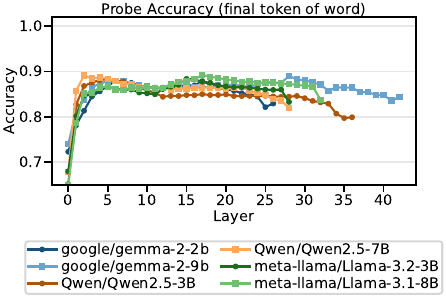}
        \caption{Across models at final token}
        \label{fig:probe-comparison}
    \end{subfigure}

    \vspace{6pt}

    \begin{subfigure}{\linewidth}
        \centering
        \includegraphics[width=\linewidth]{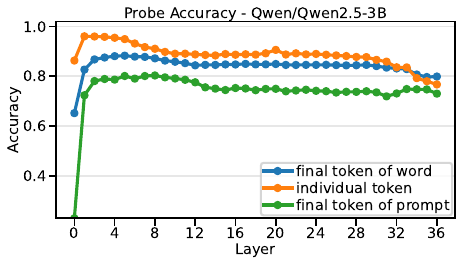}
        \caption{Across positions for Qwen2.5-3B}
        \label{fig:probe-position-comparison}
    \end{subfigure}

    \caption{Layerwise probe accuracy}
    \label{fig:probe-combined}
    \vspace{-20pt}
\end{wrapfigure}
Figure~\ref{fig:probe-comparison} compares probing performance for all the models, averaging accuracies from all the letter probes. Figure~\ref{fig:probe-position-comparison} compares probing performance for Qwen2.5-3B across all the probe positions, averaging accuracies from all the letter probes.

\textbf{Character-count information is reliably encoded.}
Probing results show that character-count information is consistently encoded in model representations as seen in Figure~\ref{fig:probe-comparison} from \textbf{high accuracies above 80\% for most layers across all the models}. Probes achieve high accuracy across all target letters on a fully held-out test set containing words and tokens never seen during training, indicating that the signal is not due to memorization. Figure~\ref{fig:probe-comparison} compares performance across models at the final token of the word and shows that letter-count information is reliably recoverable from the residual stream at this position. 

This suggests that models can both represent character-level features within tokens and compose them across subword units to form word-level counts.

\noindent\textbf{Information degrades during transfer to the prediction position.}
To understand how this information is propagated, Figure~\ref{fig:probe-position-comparison} compares probe performance at three positions for Qwen2.5-3B: individual tokens, the final token of the word, and the final token of the prompt. We observe that \textbf{token-level probes achieve the highest accuracy (higher than 90\% in the initial layers)}, indicating that models reliably encode character information within individual tokens. \textbf{Performance at the final word token remains high (reliably above 80\%)}, with only a slight drop, suggesting that models are largely able to aggregate information across subword tokens with minimal loss.

However, \textbf{probe accuracy drops substantially at the final token of the prompt (consistently below 80\%)}, which is the position used for next-token prediction. This indicates that while character-count information is present and composable within the word, it is not always completely preserved or routed to the final prediction position. The results suggest moderate information loss during this transfer, highlighting a disconnect between representation and utilization.

\noindent\textbf{Encoding is independent of final prediction correctness.}
Importantly, this encoding is not contingent on whether the model 
ultimately produces the correct answer. We observe that character-count 
information is reliably present in intermediate representations regardless of whether the final prediction is correct or incorrect. Even in cases where the model outputs the wrong count, the correct count remains decodable from the residual stream at earlier layers. This further reinforces our central finding: the failure is not one of representation, but of utilization. The model encodes the answer, but fails to carry it through to the output.

A breakdown of probe accuracy by individual target letter, revealing 
variation across the alphabet, is provided in Appendix~\ref{app:probe_heatmap}.


\subsection{Logit Lens Analysis}

\begin{figure*}[!htp]
    \centering
    \begin{subfigure}[b]{0.48\textwidth}
        \centering
        \includegraphics[width=\linewidth,trim={0 2mm 0 2mm},clip]{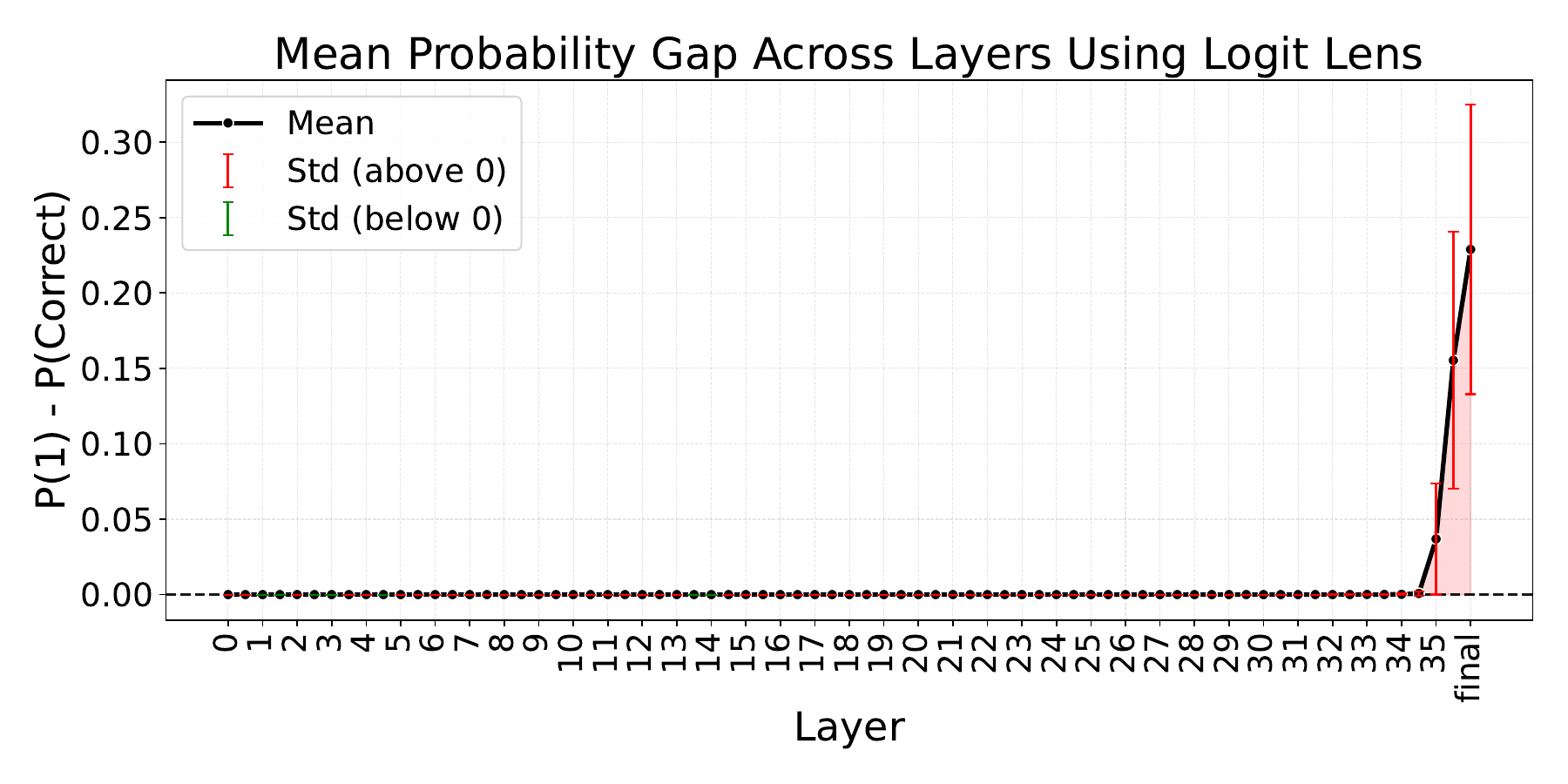}
        \caption{Qwen2.5-3B}
        \label{fig:logit-lens}
    \end{subfigure}
    \hfill
    \begin{subfigure}[b]{0.48\textwidth}
        \centering
        \includegraphics[width=\linewidth,trim={0 2mm 0 2mm},clip]{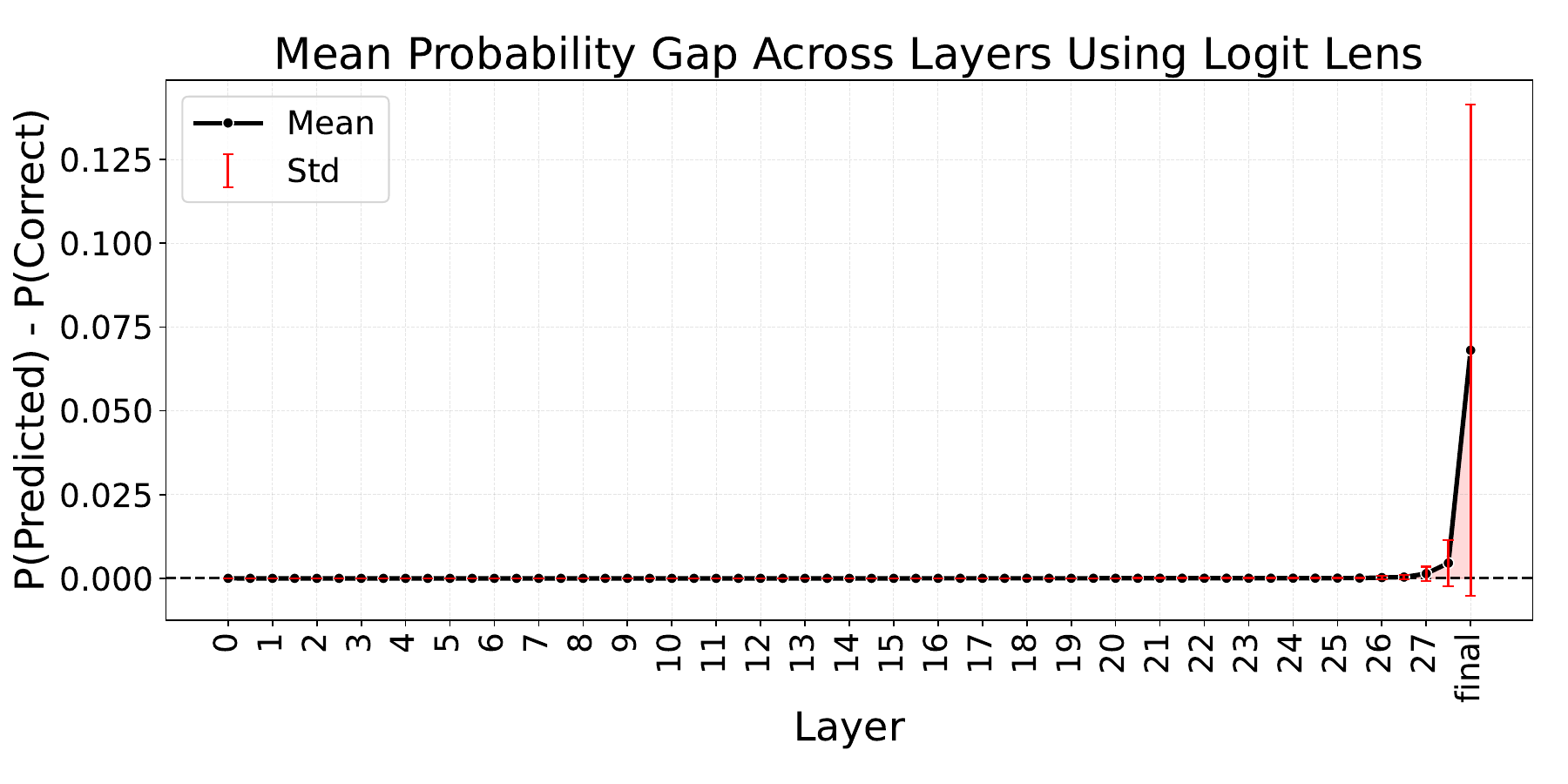}
        \caption{Llama3.2-3B}
        \label{fig:logit-lens-llama}
    \end{subfigure}
    \caption{Logit difference across layers for the correct count token, averaged over approximately 100 samples. Higher values indicate layers where suppression of the correct answer occurs. The final label \textit{final} denotes the final hidden state after all layers. Results are computed over samples where the model produces an incorrect prediction, corresponding to the ``always predict 1'' strategy for Qwen2.5-3B and the near-uniform random strategy for LLaMA3.2-3B.}
    \label{fig:logit-lens-comparison}
\end{figure*}

\noindent\textbf{Final-layer components suppress the correct answer.}
To trace how model predictions evolve across layers, we apply the logit lens and track the probability gap ($\Delta_p$), which we take as a proxy for suppression since it captures both an increase in the probability of incorrect answers and a decrease in the probability of the correct answer. Despite probes indicating that character-count information is available, Figure~\ref{fig:logit-lens-comparison} shows that later layers increasingly suppress the correct answer. We call the components comprising these layers negative components, and the circuit comprised of these negative components a negative circuit. This indicates that while the correct count is encoded in the residual stream, it is not successfully transferred from the representational space into the probability space that is ultimately consumed by the unembedding layer to produce output logits. The information exists within the model, but is not surfaced in a form that influences the final prediction.

W\textbf{e find that the final layer, particularly the MLP component, is primarily responsible for this behavior.} Instead of utilizing the available information, it amplifies degenerate answer strategies. For Qwen2.5-3B, this corresponds to the ``always predict 1'' strategy, while for LLaMA3.2-3B it corresponds to a near-uniform (random) prediction pattern.

Quantitatively, \textbf{the final layer increases the probability of the strategy-consistent answer by approximately 30\% for Qwen and 7\% on average for LLaMA.} Although variance across samples is high, the one-standard-deviation range remains predominantly in the positive regime, indicating that suppression of the correct answer is the dominant behavior.

These results suggest that the failure is not due to absence of information, but rather due to late-stage transformations that override correct signals in favor of simpler heuristics.

\noindent\textbf{Suppression is concentrated in late MLP and attention components.}
A more fine-grained analysis reveals that suppression is not uniformly distributed across the network, but is concentrated in a small set of late-layer components. \textbf{In Qwen2.5-3B, the penultimate-layer MLP suppresses approximately 4\% of probability mass on average, the final-layer attention heads contribute a further 9\% suppression, and the final-layer MLP accounts for an additional 8\%.} Taken together, the last two layers dominate the overall suppression effect.

In contrast, LLaMA3.2-3B exhibits a more localized pattern. \textbf{The penultimate-layer MLP has negligible impact, and the final-layer attention heads contribute only around 0.2\% suppression.} The effect is instead driven almost entirely by \textbf{the final-layer MLP, which suppresses approximately 7.5\% of the correct-answer probability mass.} This difference aligns with the observed behavioral patterns, where Qwen exhibits a strong deterministic bias (``predict 1''), whereas LLaMA produces near-uniform outputs.

Overall, these findings indicate that a small number of late-stage components are sufficient to override otherwise correct intermediate representations.

\subsection{Activation Patching}

\noindent\textbf{No specific Attention Heads identified in routing character information.}
From Activation patch results (Qwen2.5-3B attached in Figure \ref{fig:act-patch-attention}, the remaining attached in Appendix \ref{sec:appendix-qwen-actpatch}, \ref{sec:appendix-gemma2-actpatch}, and \ref{sec:appendix-llama-actpatch})  we observe that while information is routed from the word and letter to the subsequent tokens from the middle layers till the later layers, no single attention layer consists of an attention head that shows a strong or consistent restoration effect across layers. Instead, \textbf{mild restoration (less than 5\%) appears scattered across many different layers}. This suggests that the model does not rely on a dedicated attention layer to transfer character-level information. Additionally, we also run patching experiments on individual head outputs, and find that there are no attention heads with interpretable attention patterns that specialize in routing from the letter or word to the final token as shown in Figure \ref{fig:act-patch-attention}. \textbf{We investigate attention patterns for heads with performance restoration $>= 0.3$ and find no interpretable patterns.} We find that the information is fragmented across multiple heads, which may lead to weak or incomplete aggregation at the final token.

\begin{figure*}[!htbp]
    \centering
    \includegraphics[width=\linewidth]{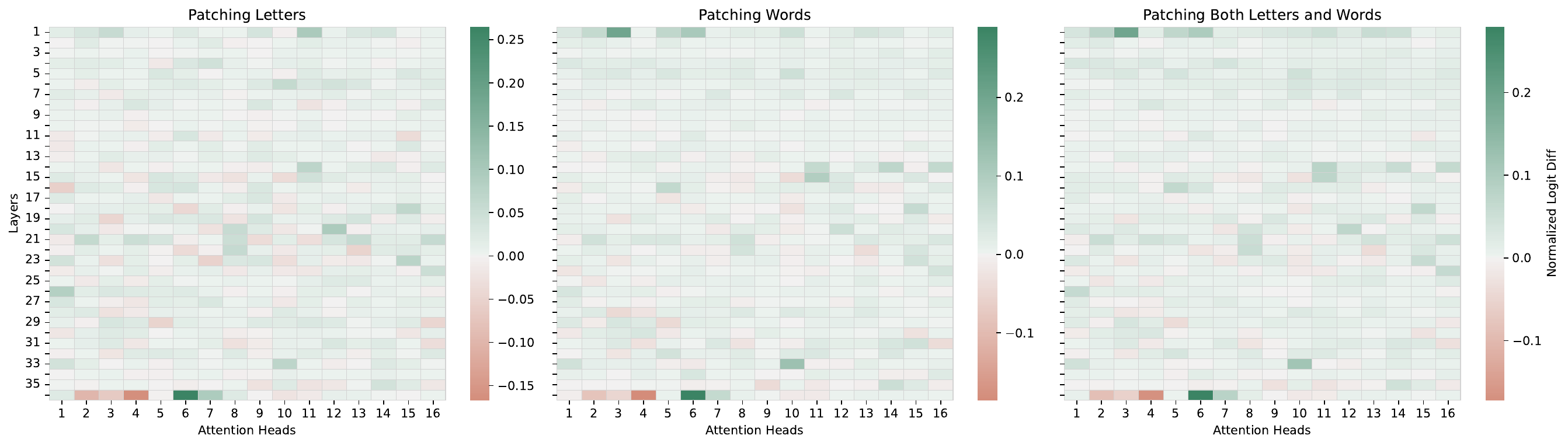}
    \caption{Activation patching heatmaps when corrupting the word, letter or both letter and word for Qwen2.5-3B attention heads averaged over 1000 samples. Green regions indicate components where clean activations restore the correct behavior, helping identify localized circuits for character counting. The Red regions indicate components which diminish performance when patched in.}
    \label{fig:act-patch-attention}
\end{figure*}

\noindent\textbf{Activation patching supports late-layer suppression.}
These activation patching results also provide further evidence for the suppressive components identified via the logit lens. While activation patching is less precise at isolating suppression than logit lens analysis, traces of suppressive behavior are still visible. \textbf{In particular, for Qwen2.5-3B, head 4 of the final layer (L32 H4) exhibits a consistently negative restoration effect in Figure~\ref{fig:act-patch-attention} (with a performance restoration around 10 - 15\%)}, indicating that patching in clean activations at this head worsens performance rather than improving it. This is consistent with our logit lens finding that the final-layer attention heads contribute approximately 9\% suppression of the correct-answer probability mass, and lends additional mechanistic support to the existence of suppressive components concentrated in the final layers.

\subsection{Increasing Scale or Post-Training Does Not Help}

We extend our experiments to larger models in the 7--9B parameter range 
and to instruction-tuned variants of all model families using prompts from 
Appendix~\ref{sec:prompt-for-IT}. We observe the same patterns, if 
not stronger, across all settings.

Scaling up by approximately 4$\times$ in parameter count does not alleviate 
the identified failure modes. The logit lens profiles, suppression patterns, 
and degenerate answer strategies remain consistent between the smaller and 
larger variants of each model family. Similarly, post-training through 
instruction tuning does not resolve the issue: instruct models continue to 
suppress the correct answer in late layers and default to a similar set of degenerate 
strategies as their base counterparts. \textbf{A similar picture emerges for 
instruction-tuned models (Appendix~\ref{app:instruct_performance}): 
post-training shifts the answer strategies adopted by most models, with 
the majority moving toward alternating between counts of ``1'' and ``2'', 
but accuracy remains close to random. (around 40\%) }Qwen2.5-3B-Instruct is the only 
exception, retaining its ``always predict 1'' strategy and achieving 
marginally higher accuracy as a result, though this reflects a more 
favorable strategy rather than any genuine improvement in counting ability.

Logit lens analysis on the larger and instruction-tuned variants reveals 
a very similar suppression profile to that of the smaller base models. 
The same concentration of suppressive behavior in the penultimate and 
final layer MLP and attention components is consistently observed, 
confirming that the suppression mechanism is not an artifact of model 
size or training regime but a structural property of how these models 
process character-level information.

This is particularly striking because our probing results show that even 
the smaller models in our study reliably encode character-count information 
in their intermediate representations. Larger models can only be expected 
to encode such low-level orthographic features more strongly, yet the 
suppression at later layers persists regardless. The bottleneck is not 
representational capacity but the learned computation that overrides correct 
signals at the output stage. The model encodes the answer, but has learned 
to instead commit to a simpler answer strategy during generation.

This observation is consistent with reports that even the largest 
proprietary models, such as GPT-4, struggle with character-level counting 
tasks without the aid of chain-of-thought prompting, tool use, or other 
reasoning scaffolds. If 
scaling alone were sufficient to fix these failures, such tricks would not 
be necessary. Our mechanistic findings suggest a reason: the failure is 
structural, arising from suppressive  components employing degenerate answer strategies that are reinforced rather 
than eliminated by scale and post-training.

\section{Related Work}

Prior work has introduced benchmarks targeting symbolic and character-level reasoning, noting similar deficiencies even in priopreitray models like GPT-4 \citep{achiam2023gpt} and LLaMA‑3 \citep{dubey2024llama} still underperform on tasks such as counting and position-specific edits \citep{zhang2024large}.
LMentry \citep{efrat2022lmentry} proposes minimal tasks such as letter containment and first/last character identification, while CUTE \citep{edman2024cute} expands this framework to include spelling, character edits, and orthographic similarity—highlighting persistent failures even in instruction-tuned models. Building on these efforts, CWUM \citep{zhang2024large} provides a comprehensive suite of fifteen character- and word-level manipulation tasks to systematically evaluate these limitations.
Prior studies have noted similar deficiencies in basic arithmetic and symbolic reasoning (Shin et al., 2024; Hanna et al., 2023), but the mechanisms behind these failures remain poorly understood.

\section{Discussion \& Conclusions}
Our results show that models like LLaMA, Qwen, and Gemma handle character-level information in surprisingly inconsistent ways. Even when these models seem to represent letters quite well in their early layers, they often fail to carry that information forward to later computations. This pattern suggests that the problem is not with the model's capacity to encode such information, but rather with how they pass it through different parts of the network.

One striking observation is the lack of any clear attention heads or components dedicated to this kind of processing. Instead, the signal appears scattered and weakly coordinated across many layers. In practice, this means that character-level reasoning may not be an explicit feature of these systems, it emerges only as a side effect of training, rather than as something the model is designed to handle well.

This insight also connects to a broader question about what kind of linguistic structure large language models truly capture. If they struggle to count letters correctly, it raises doubts about their ability to handle other tasks that depend on fine-grained symbolic precision, such as spelling correction, transliteration, or reasoning over structured text. Future research might need to explore whether additional training objectives or architectural changes could strengthen these low-level capabilities without compromising the higher-level semantics that make these models so powerful.

\section{Limitations}
While this study provides useful insights, it also has several important limitations. First, we only analyzed models in the 2–3B and 7–9B parameter range. These sizes make interpretability experiments more manageable, but they don’t tell us whether the same behavior appears in much larger systems like GPT-4~\citep{openai2024gpt4technicalreport} or Mistral~\citep{jiang2023mistral7b}.
Second, our evaluation setup uses single-word, controlled inputs, which help isolate specific effects but do not capture the complexity of full sentences or natural text.
Third, all experiments were conducted in English, meaning our findings may not generalize to languages with richer morphology or non-alphabetic scripts. Fourth, while methods such as activation patching and probing give a useful window into model internals, they don’t guarantee a complete causal picture. The true computation might involve more subtle, distributed interactions that these tools can’t fully capture.
Finally, our work focuses on identifying failure modes rather than fixing them. We have not yet tested whether fine-tuning, additional supervision, or architectural adjustments could reduce these errors. Exploring such interventions would be a valuable direction for future work.

\section{Ethical Considerations}
This work focuses on analyzing the internal behavior of large language models (LLMs) through controlled diagnostic tasks. All experiments are performed using open-access models and publicly available datasets, and no sensitive, personal, or private data is used at any stage. The benchmark tasks are synthetic by design and do not involve any real-world user data. The goal of this study is to enhance understanding and transparency in how large models process low-level linguistic information, particularly in cases where they fail on tasks that appear trivial to humans. By identifying such systematic failure modes, we aim to contribute to the responsible development and deployment of LLMs, offering insights that may inform future improvements in architecture, training, and evaluation. This work does not involve human participants, crowdsourced annotations, or generation of sensitive content, and therefore presents minimal ethical or societal risk.

\bibliographystyle{colm2026_conference}
\bibliography{custom}

\newpage
\appendix
{\Large\textbf{Appendix}}

\section{On the Inapplicability of Control Tasks to Our Probing Setting}
\label{app:control_tasks}

\citet{hewitt-liang-2019-designing} propose \textit{control tasks} as a 
diagnostic for probing experiments: by associating word types with random 
outputs, they measure whether a probe succeeds by genuinely extracting 
information from representations, or merely by memorizing word-type-to-label 
mappings. The resulting selectivity metric is the gap between linguistic task 
accuracy and control task accuracy, and is intended to guard against probes 
that solve tasks through memorization rather than representation decoding. 
While the validity of this framework has itself been questioned 
\citep{pimentel-etal-2020-information}, we argue on principled grounds that 
control tasks are ill-suited to our setting, and that the concern they 
address simply does not arise here.

The core worry behind control tasks is that a probe might learn to perform 
a linguistic task such as POS tagging or dependency relation prediction by 
exploiting the correlation between word identity and the typical label 
associated with that word type. This failure mode is plausible precisely 
because such tasks admit memorizable surface-form-to-label associations.

Our task has a fundamentally different structure. Each probe is trained 
for a specific target letter, meaning the mapping from word type to count 
is deterministic and fixed for a given probe. However, our dataset 
construction explicitly ensures that no word or token type seen during 
probe training appears in the validation or test sets. This type-disjoint 
split means that even if a probe attempted to memorize word-type-to-count 
associations, it would have no basis to do so for unseen types, and would 
achieve chance performance on the test set. The concern motivating control 
tasks is that a probe might succeed by memorizing surface-form-to-label 
mappings rather than decoding the representation. This concern is 
structurally eliminated by our dataset design, following the approach 
advocated by \citet{hupkes-etal-2018-visualisation} and acknowledged as 
a valid mitigation strategy by \citet{hewitt-liang-2019-designing} themselves.

This guarantee is especially robust for multi-token words. When a word is 
split into multiple subword tokens, the correct character count depends on 
the specific combination of tokens and the target letter, a combination 
that is not repeated across splits. Novel words at test time present 
entirely new token combinations with counts never encountered during 
training, making memorization structurally impossible.

A further property that distinguishes our setting is that the correct output 
cannot be inferred from token identifiers alone. Character count is an 
orthographic property of the word's surface form, information that is not 
directly encoded in token IDs and cannot be recovered without access to the 
actual character sequence. A probe operating purely on token identity, 
without genuinely decoding character-level features from the hidden 
representation, has no basis on which to perform the task.

Finally, we use the simplest possible probe architecture: a single linear 
layer without bias (Equation~\ref{eq:probe}), which has the lowest possible 
capacity for memorization and is known to yield high selectivity even in 
standard probing settings \citep{hewitt-liang-2019-designing}.

Together, these properties, namely type-disjoint splits, orthographic 
dependence, and minimal probe complexity, ensure that our probing results 
reflect genuine representation content rather than superficial memorization.

\section{Connection Between Cross-Entropy Loss and Logit Difference}
\label{app:cross_entropy}

The logit difference metric $\Delta$ is closely related to the standard 
cross-entropy loss used during training. Specifically, taking differences 
in cross-entropy loss between two tokens is equivalent to taking their 
logit difference, as the softmax normalizer cancels out.

Recall that the cross-entropy loss for a token $a$ is:

\begin{equation}
    -\log P(a) = -\log\left(\frac{e^{\text{logit}(a)}}{\sum_{v \in V} e^{\text{logit}(v)}}\right)
\end{equation}

Expanding the logarithm:

\begin{equation}
    -\log P(a) = -\log\left(e^{\text{logit}(a)}\right) + \log\left(\sum_{v \in V} e^{\text{logit}(v)}\right)
\end{equation}

\begin{equation}
    -\log P(a) = -\left(\text{logit}(a) - \log\left(\sum_{v \in V} e^{\text{logit}(v)}\right)\right)
\end{equation}

Now consider the difference in cross-entropy loss between two tokens 
$a$ and $b$:

\begin{equation}
    -\log P(a) - \left(-\log P(b)\right) = -\text{logit}(a) + \text{logit}(b)
\end{equation}

The log-sum-exp term $\log\left(\sum_{v \in V} e^{\text{logit}(v)}\right)$ 
cancels exactly, leaving a pure logit difference. Consequently, $\Delta > 0$ 
corresponds precisely to the correct token having a lower cross-entropy 
loss than the competing token, providing a direct link between our 
interpretability metric and the training objective.

\section{Confusion Matrices and Answer Strategies}
\label{app:confusion_matrices}

Inspecting the confusion matrices of model predictions reveals two distinct 
degenerate answer strategies that account for the poor overall performance 
observed in Section~\ref{sec:overall_perf}.

\begin{figure}[h]
    \centering
    \begin{subfigure}[t]{0.48\textwidth}
        \centering
        \includegraphics[width=\textwidth]{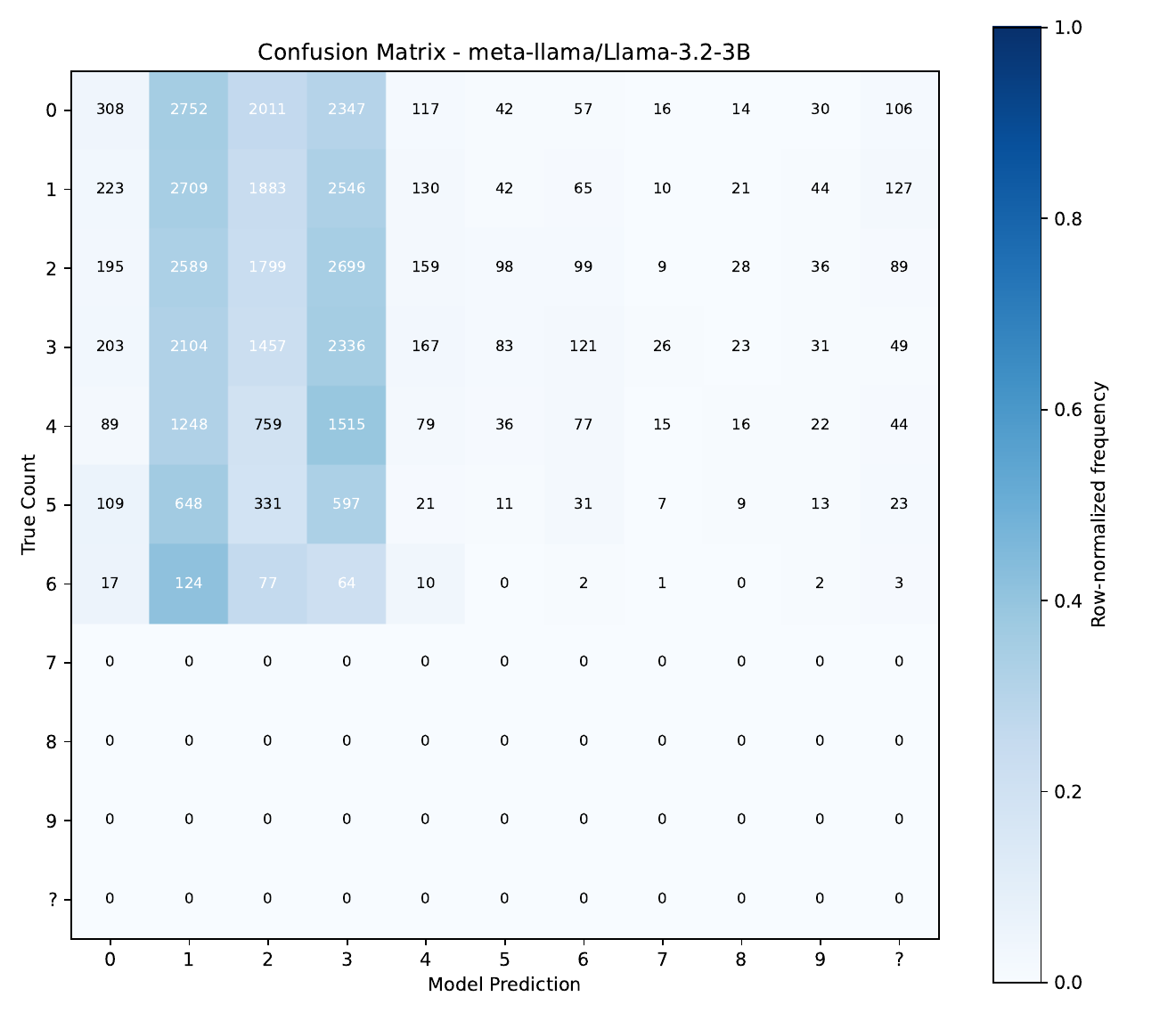}
        \caption{LLaMA-3.2-3B exhibits a near-uniform prediction distribution 
        across all count values, corresponding to a effectively random answer 
        strategy that is agnostic to the input.}
        \label{fig:confusion_llama}
    \end{subfigure}
    \hfill
    \begin{subfigure}[t]{0.48\textwidth}
        \centering
        \includegraphics[width=\textwidth]{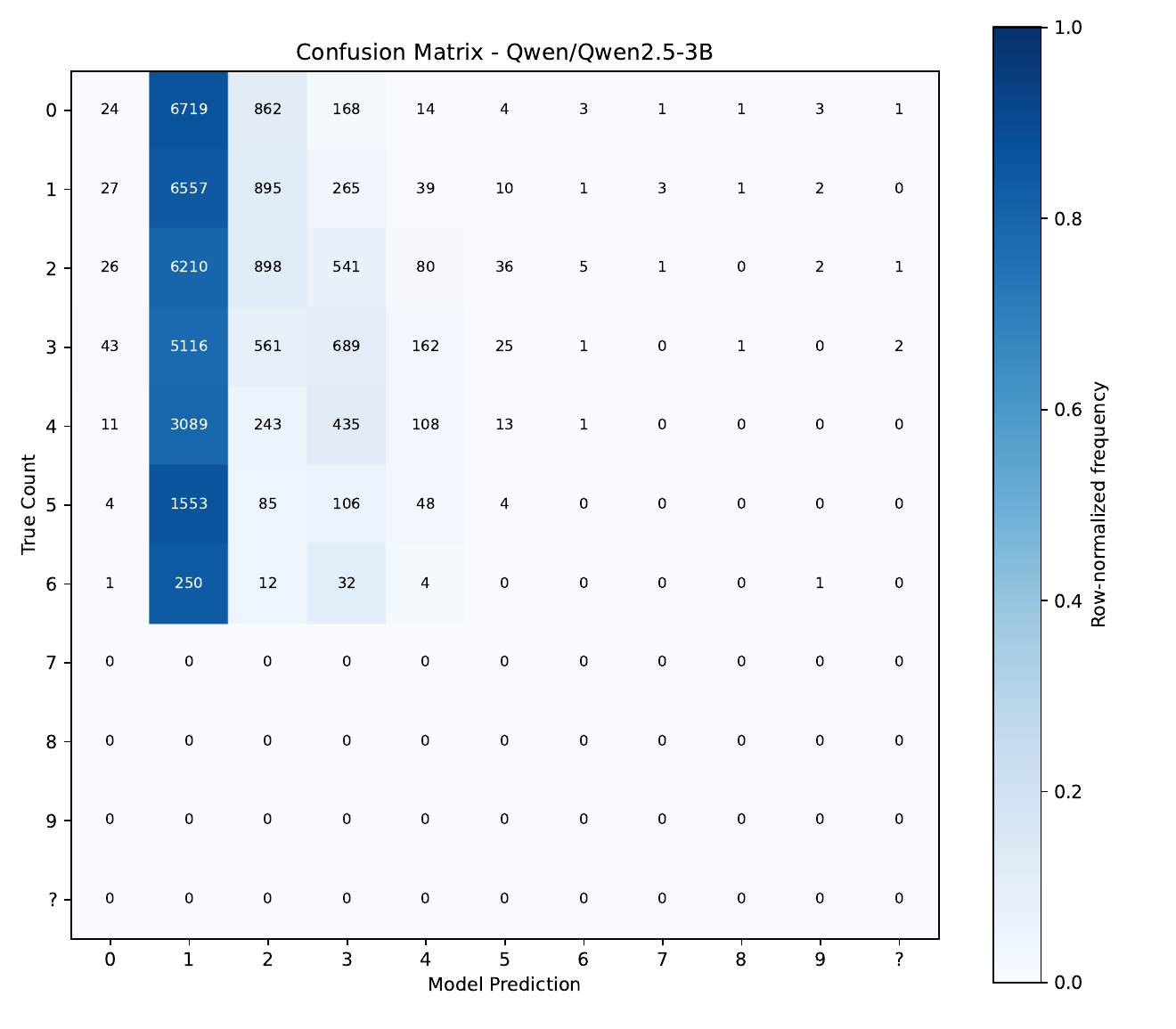}
        \caption{Qwen2.5-3B overwhelmingly predicts the count ``1'' regardless 
        of the true count, corresponding to a degenerate constant prediction 
        strategy.}
        \label{fig:confusion_qwen}
    \end{subfigure}
    \caption{Confusion matrices for predicted versus true letter counts. The 
    two models represent the two dominant failure modes observed across all 
    model families: a random strategy (LLaMA) and a constant prediction 
    strategy (Qwen).}
    \label{fig:confusion_matrices}
\end{figure}

The two strategies reflect qualitatively different failure modes. LLaMA-3.2-3B 
distributes its predictions roughly uniformly across possible counts, 
suggesting that the model does not develop any stable preference and 
effectively ignores the input when producing an answer. This explains 
why its accuracy sits close to the random baseline in Figure~\ref{fig:perf_plot}. 
Qwen2.5-3B, by contrast, adopts a highly deterministic but incorrect 
strategy, defaulting to predicting a count of ``1'' in the vast majority 
of cases. This reflects a strong learned prior toward low counts, likely 
reinforced during training on natural language where single letter 
occurrences are most common. Gemma-2 models exhibit behavior qualitatively 
similar to Qwen, also defaulting to constant low-count predictions. These 
observations confirm that the models are not attempting to perform genuine 
counting, but are instead falling back on simple heuristics, consistent 
with the suppression of correct intermediate representations identified 
in our mechanistic analysis.

\section{Performance Analysis}
\label{app:perf_analysis}

\begin{figure}[t]
    \centering
   \includegraphics[width=\linewidth,trim={0 2mm 0 2mm},clip]{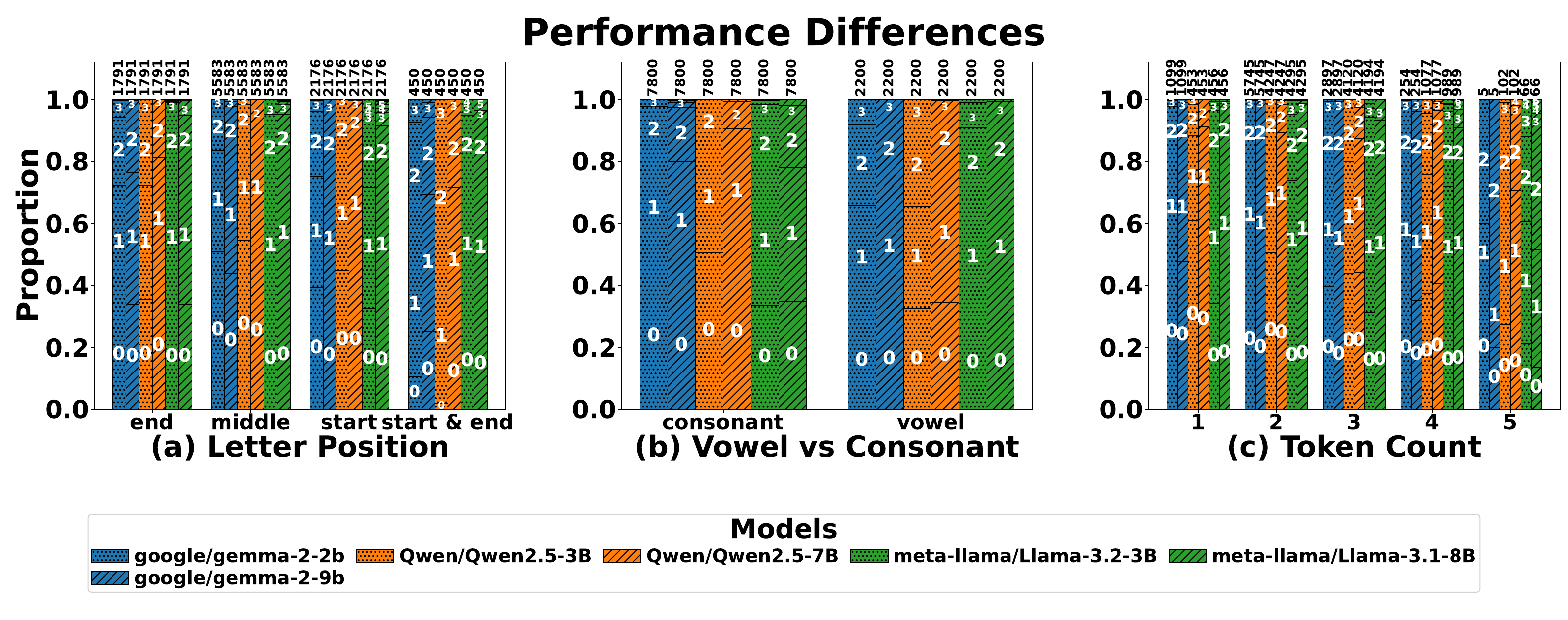}
    \caption{Distributions of absolute deviations between the correct letter count and the predicted count ($|\text{predicted count} - \text{correct count}|$). The y-axis indicates the proportion of predictions at each deviation. Every stacked bar represents a distribution of deviations. Bars labelled with values above zero reflect over- or under-counting errors relative to the true letter count and the bars labelled $\text{deviation} = 0$ represent the accuracy (no deviation). The number on top of every bar denotes how many instances of the category were present.}
    \label{fig:perf-analysis}
\end{figure}

We perform a performance analysis across different parameters. We find the following effects:

\paragraph{Effect of Letter Position}
We evaluate how performance varies depending on whether the target letter appears at the start, end, or both ends of a word, compared to when it occurs in the middle. As shown in Figure~\ref{fig:perf-analysis} (a), models consistently struggle more with letters at the extremes. This difficulty is reflected in the higher proportion of zero-correct predictions (bars labeled 0) for start, end, and start–end positions. The performance drop is especially noticeable for Qwen (blue) and Gemma (orange), whereas LLaMA remains close to random performance across all positions.

\paragraph{Vowel vs. Consonant Counting}
We compare model performance on counting vowels versus consonants. As shown in Figure~\ref{fig:perf-analysis} (b), models generally struggle more with vowel counting. This is evident from the higher proportion of zero-correct predictions (bars labeled 0) in the vowel condition. The performance drop is particularly for Qwen (blue) and Gemma (orange), while LLaMA shows consistently low accuracy, remaining close to random performance.

\paragraph{Token Counts}
We also examine how performance varies with the number of tokens that compose a single word. As shown in Figure~\ref{fig:perf-analysis} (c), models tend to perform better when words are represented by fewer subword tokens. This trend is evident from the decline in accuracy (bars labelled 0) as token count increases. In other words, when a word is split into more subword units, models struggle more with accurately counting its letters.

\section{Per-Letter Probe Accuracy Heatmaps}
\label{app:probe_heatmap}

The probing results presented in Section~\ref{sec:probing} average 
accuracy across all target letters. Here we present the full per-letter 
breakdown as heatmaps, where each row corresponds to a layer and each 
column corresponds to a specific target letter. This allows us to inspect 
whether certain letters are encoded more reliably than others, and whether 
the layer at which encoding peaks varies across letters.

\begin{figure}[h]
    \centering
    \includegraphics[width=\textwidth]{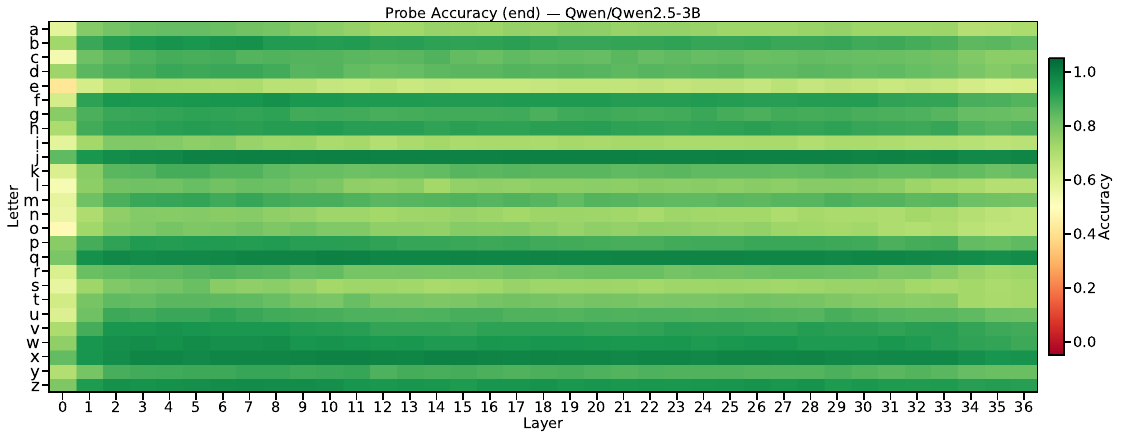}
    \caption{Layerwise probe accuracy heatmap for individual target letters 
    at the final token of the word for Qwen2.5-3B. Each cell represents 
    the probe accuracy for a specific layer and target letter.}
    \label{fig:heatmap-qwen}
\end{figure}

Overall, the heatmaps confirm that the high average probe accuracy 
reported in the main text is not driven by a small subset of easy letters. 
Character-count information is broadly and consistently encoded across the 
alphabet, with most letters achieving high accuracy by the mid layers and 
maintaining it through to the final token of the word.

\section{Prompts for Instruction-tuned Models}
\label{sec:prompt-for-IT}

We construct User and Assistant Turns for Instruction Tuned Models. We use the model's chat templates to get properly formatted text (Including the default system prompts). We show the prompts in Figure~\ref{fig:char_count_prompts}.

\begin{figure*}[!htbp]
\begin{tcolorbox}[colback=blue!5!white, colframe=blue!75!black, 
title=Character Counting Prompts, label=box:char_count_prompts, width=\textwidth]
We used the following natural language prompts to query the models for character counting. 
Each prompt contains placeholders \texttt{<letter>} and \texttt{<count\_subject>}, 
which were replaced with the target character and input string respectively.
\\[4pt]

\begin{enumerate}
    \item How many \texttt{<letter>}'s are in \texttt{<count\_subject>}?\\
          \textbf{Expected response:} The number is \texttt{<count>}.
    \item Can you count the letter \texttt{<letter>} in \texttt{<count\_subject>}?\\
          \textbf{Expected response:} The count is \texttt{<count>}.
    \item Tell me how many times \texttt{<letter>} appears in \texttt{<count\_subject>}.\\
          \textbf{Expected response:} The number is \texttt{<count>}.
    \item What is the count of \texttt{<letter>} in \texttt{<count\_subject>}?\\
          \textbf{Expected response:} The count is \texttt{<count>}.
    \item Please calculate the total number of \texttt{<letter>}'s in \texttt{<count\_subject>}.\\
          \textbf{Expected response:} The total number is \texttt{<count>}.
    \item Could you find out how many \texttt{<letter>}'s are in \texttt{<count\_subject>}?\\
          \textbf{Expected response:} The number is \texttt{<count>}.
    \item I want to know the frequency of the letter \texttt{<letter>} in \texttt{<count\_subject>}.\\
          \textbf{Expected response:} The frequency is \texttt{<count>}.
    \item Count all the \texttt{<letter>}'s present in \texttt{<count\_subject>}.\\
          \textbf{Expected response:} The total number is \texttt{<count>}.
    \item How often does \texttt{<letter>} appear in \texttt{<count\_subject>}?\\
          \textbf{Expected response:} The number of times is \texttt{<count>}.

\end{enumerate}
\end{tcolorbox}
\caption{Prompt templates used for the character counting task. Each template was evaluated under the same experimental setting.}
\label{fig:char_count_prompts}
\end{figure*}

\section{Overall Performance of Instruction-Tuned Models}
\label{app:instruct_performance}
Instruction-tuned models remain close to random performance on the 
character counting task, indicating that post-training does not resolve 
the underlying failure modes identified in our mechanistic analysis. 
However, inspecting the confusion matrices reveals that post-training 
does shift the answer strategies adopted by most models. Nearly all 
instruct variants abandon their base model strategy in favor of 
alternating between counts of ``1'' and ``2'' as the dominant prediction, 
suggesting that instruction tuning alters the surface behavior without 
fixing the underlying representational bottleneck.

\begin{wrapfigure}{r}{0.5\textwidth}
    \centering
    \vspace{-6pt}
    \includegraphics[width=0.48\textwidth]{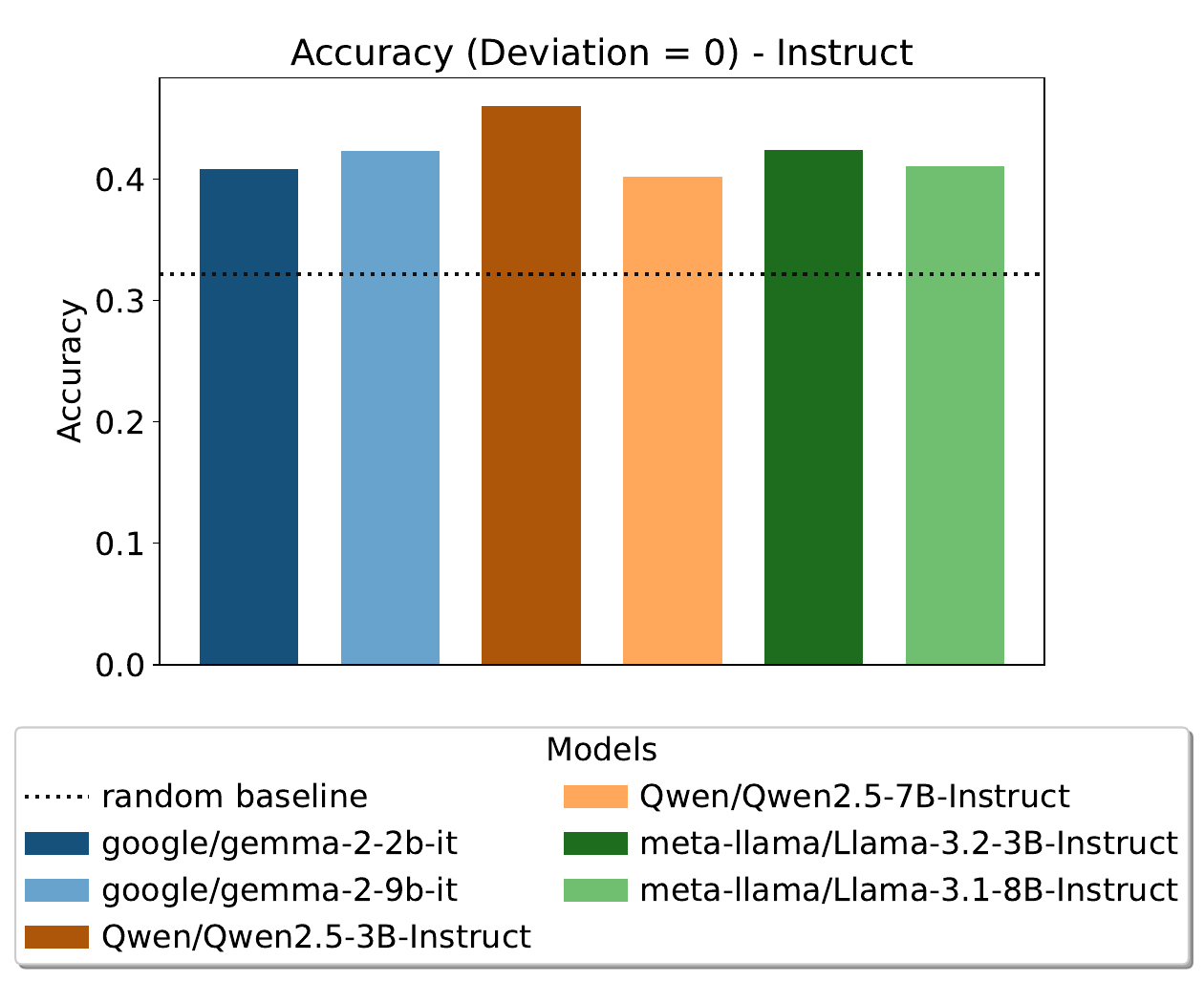}
    \caption{Accuracy of predicted letter counts for instruction-tuned 
    model variants. The dotted line indicates the random baseline accuracy 
    for comparison.}
    \label{fig:instruct-accuracy}
    \vspace{-40pt}
\end{wrapfigure}
Qwen2.5-3B-Instruct is a notable exception. It retains the ``always 
predict 1'' strategy of its base counterpart in the majority of cases, 
and as a result achieves slightly better accuracy than the other instruct 
models, since predicting ``1'' is correct more often than a uniform 
random strategy given the natural distribution of letter counts in words. 
This further illustrates that the marginal performance differences between 
models reflect differences in answer strategy rather than any genuine 
improvement in character-level reasoning.

\section{Experiments with LLaMa models}
\label{sec:appendix-llama}
To assess the generality of our observations across model scales, we replicate key experiments on LLaMA-3.1-8B, LLaMA-3.1-8B-Instruct, LLaMA-3.2-3B, and LLaMA-3.2-3B-Instruct.

\subsection{Activation Patching}
\label{sec:appendix-llama-actpatch}
\begin{figure*}[t]
    \centering
    \begin{subfigure}[a]{\textwidth}
        \centering
        \includegraphics[width=\textwidth]{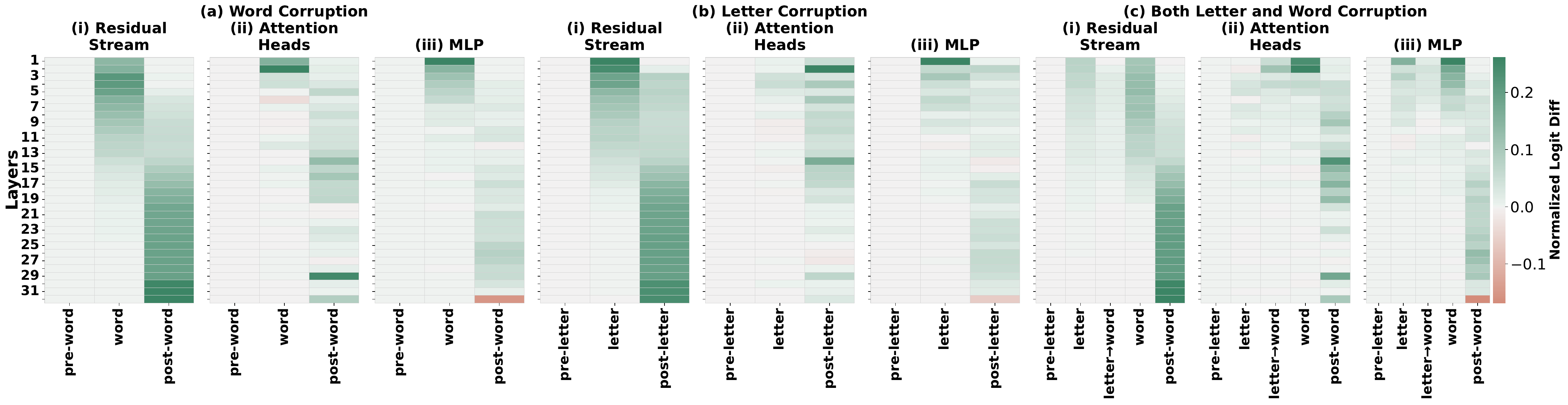}
        \caption{LLaMA-3.1-8B}
        \label{fig:llama8b-act-patch}
    \end{subfigure}
    \hfill
    \begin{subfigure}[b]{\textwidth}
        \centering
        \includegraphics[width=\textwidth]{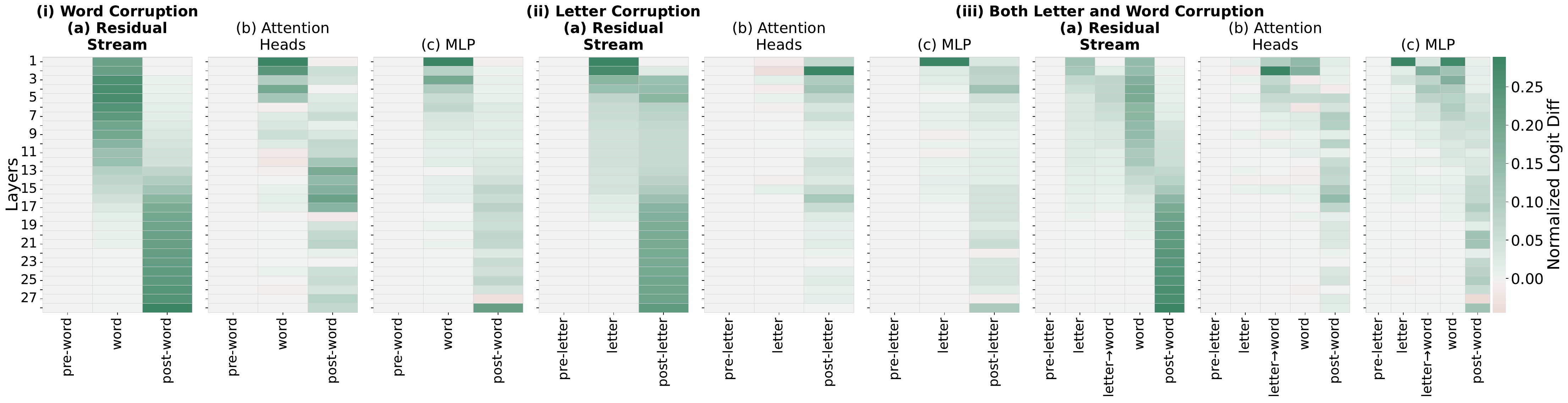}
        \caption{LLaMA-3.2-3B}
        \label{fig:llama3b-actpatch}
    \end{subfigure}
    \begin{subfigure}[c]{\textwidth}
        \centering
        \includegraphics[width=\textwidth]{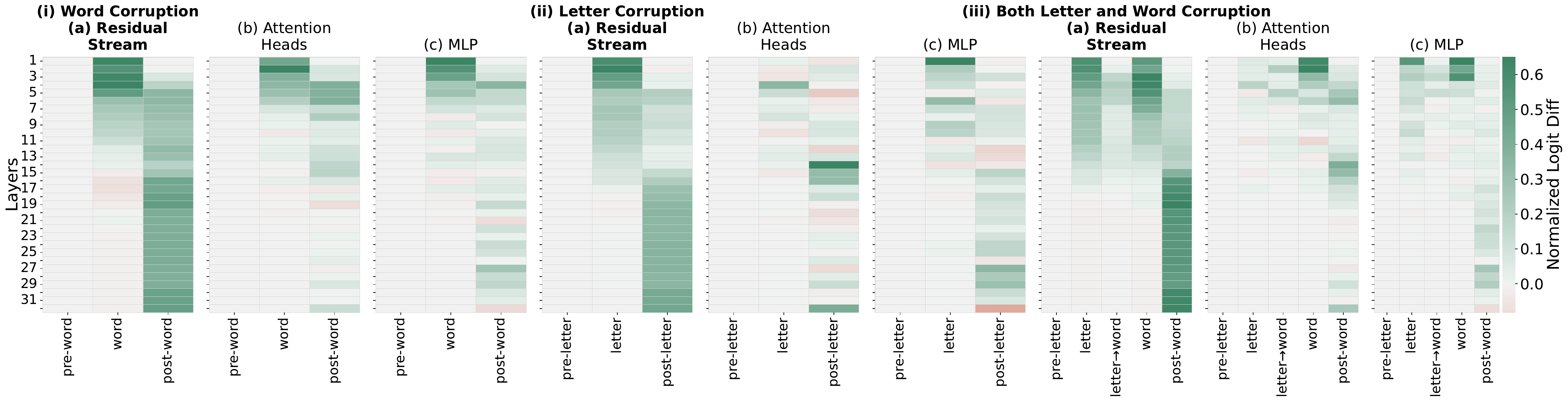}
        \caption{LLaMA-3.1-8B-Instruct}
        \label{fig:llama8b-it-actpatch}
    \end{subfigure}
    \begin{subfigure}[d]{\textwidth}
        \centering
        \includegraphics[width=\textwidth]{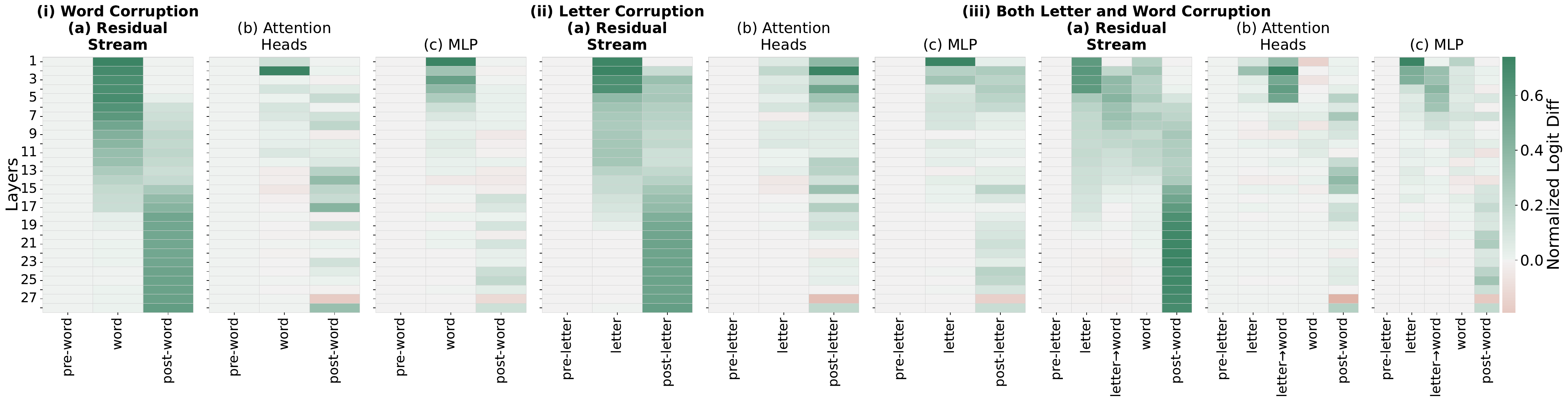}
        \caption{LLaMA-3.2-3B-Instruct}
        \label{fig:llama3b-it-actpatch}
    \end{subfigure}
    \caption{Activation patching heatmaps for Llama Models averaged over 100 samples. Green regions indicate components where clean activations restore the correct behavior, helping identify localized circuits for character counting.}
    \label{fig:llama-actpatch-combined}
\end{figure*}

\begin{figure*}[t]
    \centering
    \begin{subfigure}[a]{\textwidth}
        \centering
        \includegraphics[width=\textwidth]{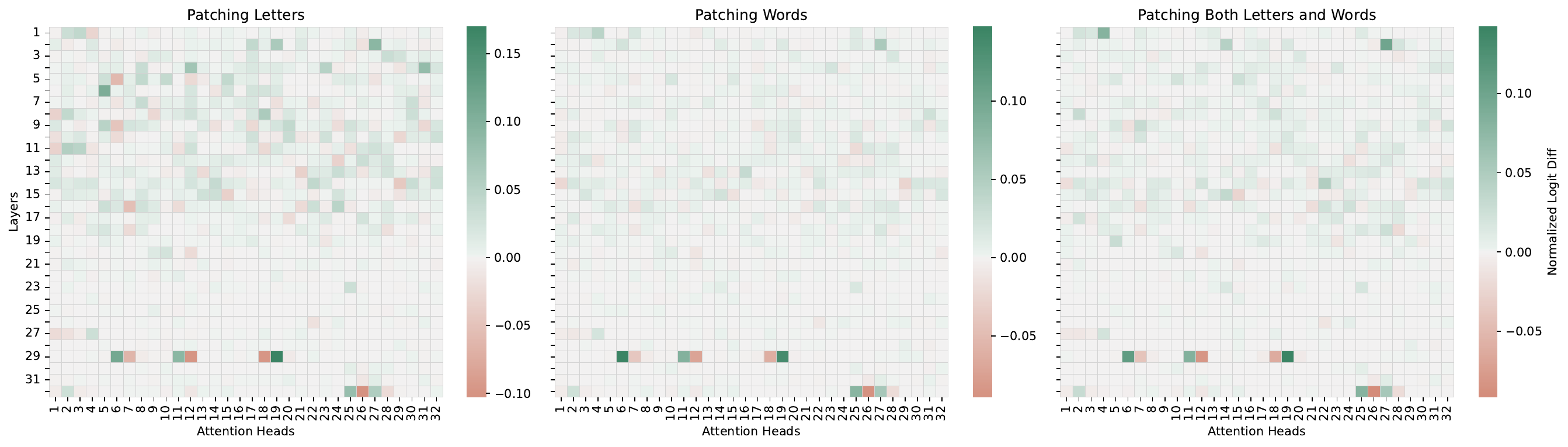}
        \caption{LLaMA-3.1-8B}
        \label{fig:llama8b-act-patch-attention}
    \end{subfigure}
    \hfill
    \begin{subfigure}[b]{\textwidth}
        \centering
        \includegraphics[width=\textwidth]{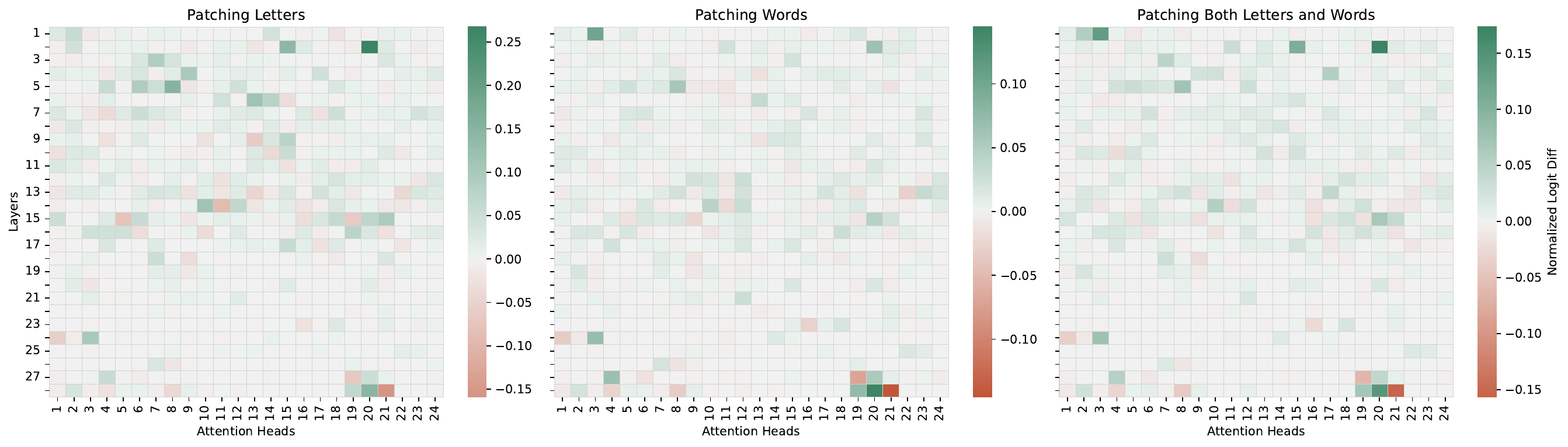}
        \caption{LLaMA-3.2-3B}
        \label{fig:llama3b-actpatch-attention}
    \end{subfigure}
    \begin{subfigure}[c]{\textwidth}
        \centering
        \includegraphics[width=\textwidth]{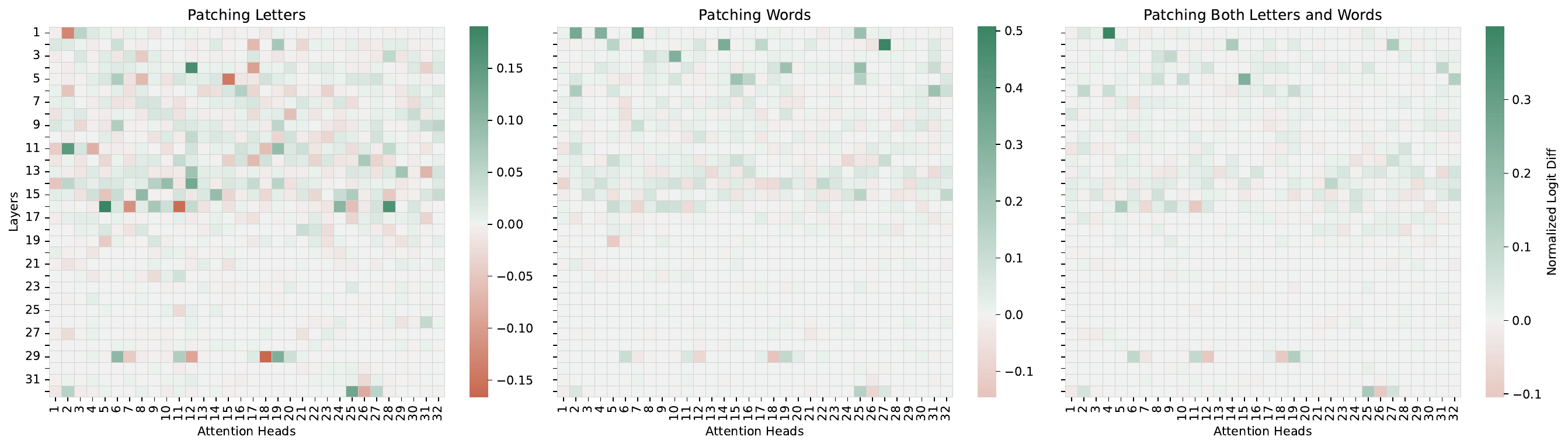}
        \caption{LLaMA-3.1-8B-Instruct}
        \label{fig:llama8b-it-actpatch-attention}
    \end{subfigure}
    \begin{subfigure}[d]{\textwidth}
        \centering
        \includegraphics[width=\textwidth]{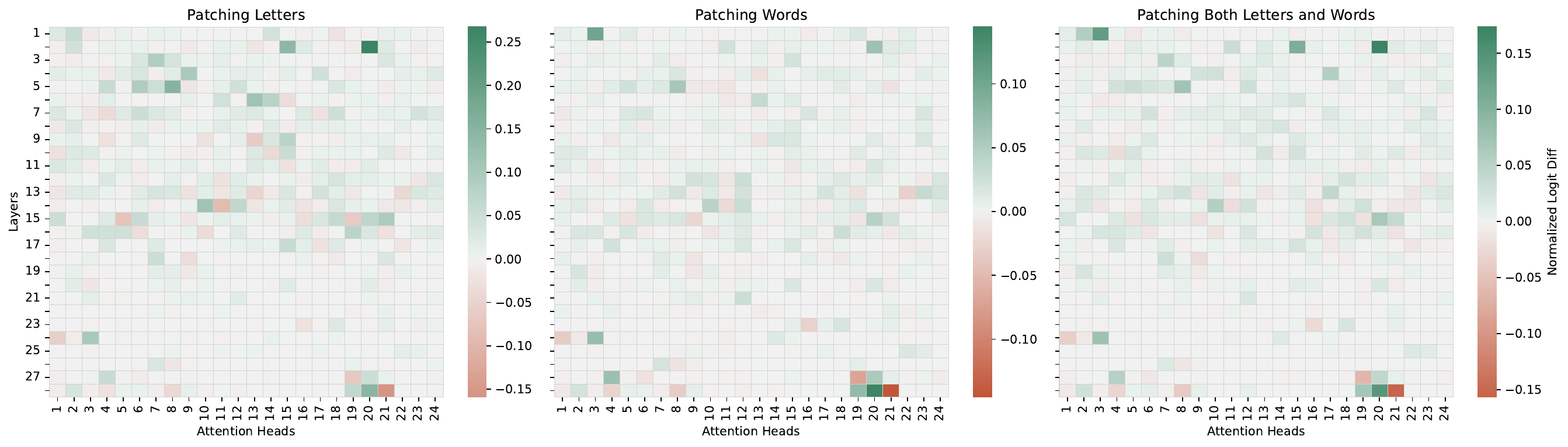}
        \caption{LLaMA-3.2-3B-Instruct}
        \label{fig:llama3b-it-actpatch-attention}
    \end{subfigure}
    \caption{Activation patching heatmaps when corrupting the word, letter or both letter and word for Llama attention heads averaged over 100 samples. Green regions indicate components where clean activations restore the correct behavior, helping identify localized circuits for character counting. The Red regions indicate components which diminish performance when patched in.}
    \label{fig:llama-actpatch-attention}
\end{figure*}

We perform activation patching as described in Section~\ref{sec:act-patch-method}. Figures~\ref{fig:llama-actpatch-combined} show restoration heatmaps for word-level, letter-level, and  both word and letter level corruptions across residual streams, attention heads, and MLP blocks for Llama Models. Figure~\ref{fig:llama-actpatch-attention} shows the restoration effects on a per-attention head basis.

\section{Experiments with Gemma-2}
\label{sec:appendix-gemma}
To examine the generality of our findings across model scales, we replicate key experiments on Gemma-2-2B and Gemma-2-9B, as well as their aligned instruction-tuned counterparts, Gemma-2-2B-IT and Gemma-2-9B-IT.

\subsection{Activation Patching}
\label{sec:appendix-gemma2-actpatch}
\begin{figure*}[t]
    \centering
    \begin{subfigure}[a]{\textwidth}
        \centering
        \includegraphics[width=\textwidth]{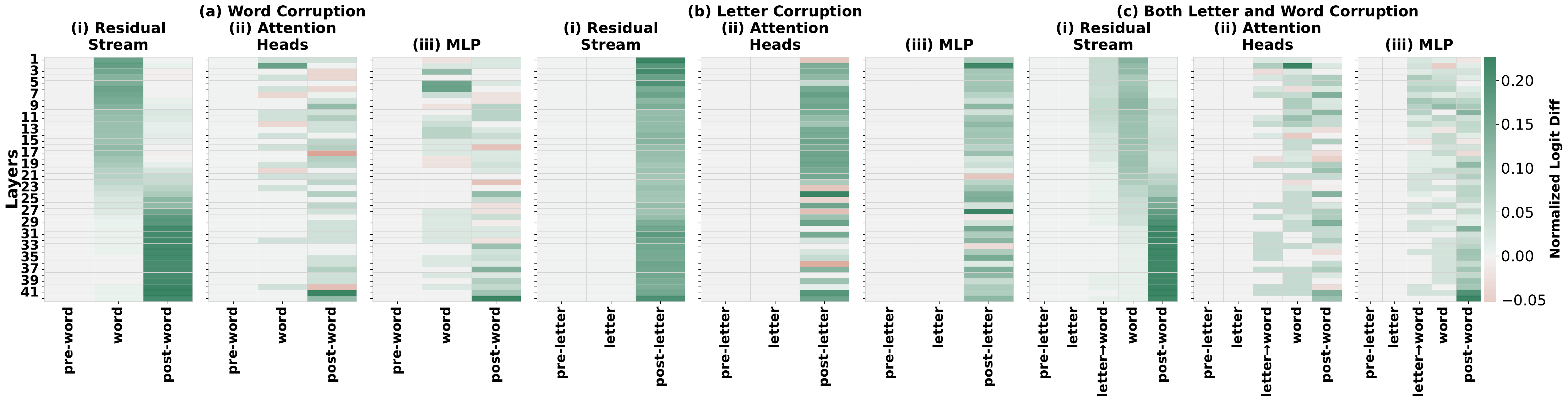}
        \caption{Gemma2-9B}
        \label{fig:gemma9b-act-patch}
    \end{subfigure}
    \hfill
    \begin{subfigure}[b]{\textwidth}
        \centering
        \includegraphics[width=\textwidth]{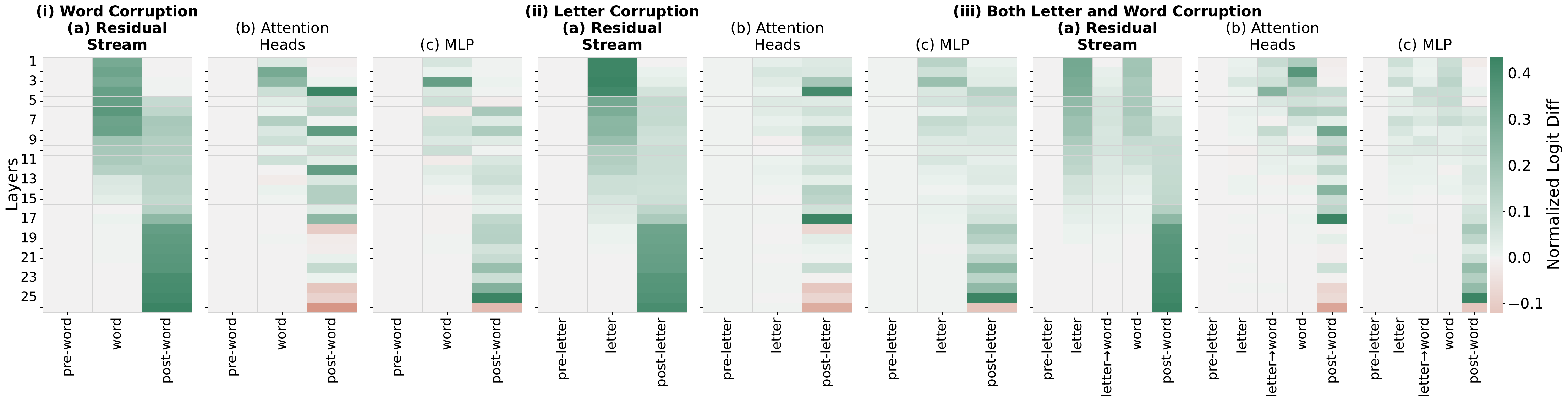}
        \caption{Gemma2-2B}
        \label{fig:gemma2b-actpatch}
    \end{subfigure}
    \begin{subfigure}[d]{\textwidth}
        \centering
        \includegraphics[width=\textwidth]{figures/Gemma2-2B-Instruct/activation_patch_combined.pdf}
        \caption{Gemma2-2B-Instruct}
        \label{fig:gemma2b-it-actpatch}
    \end{subfigure}
    \caption{Activation patching heatmaps for Gemma Models averaged over 100 samples. Green regions indicate components where clean activations restore the correct behavior, helping identify localized circuits for character counting.}
    \label{fig:gemma-actpatch-combined}
\end{figure*}

\begin{figure*}[t]
    \centering
    \begin{subfigure}[a]{\textwidth}
        \centering
        \includegraphics[width=\textwidth]{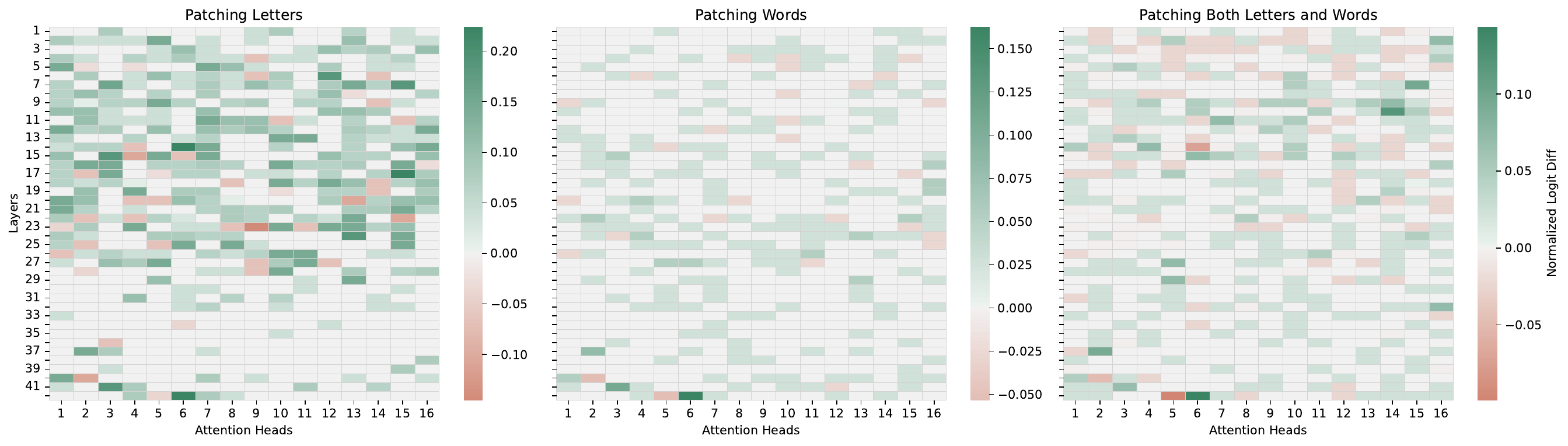}
        \caption{Gemma2-9B}
        \label{fig:gemma9b-act-patch-attention}
    \end{subfigure}
    \hfill
    \begin{subfigure}[b]{\textwidth}
        \centering
        \includegraphics[width=\textwidth]{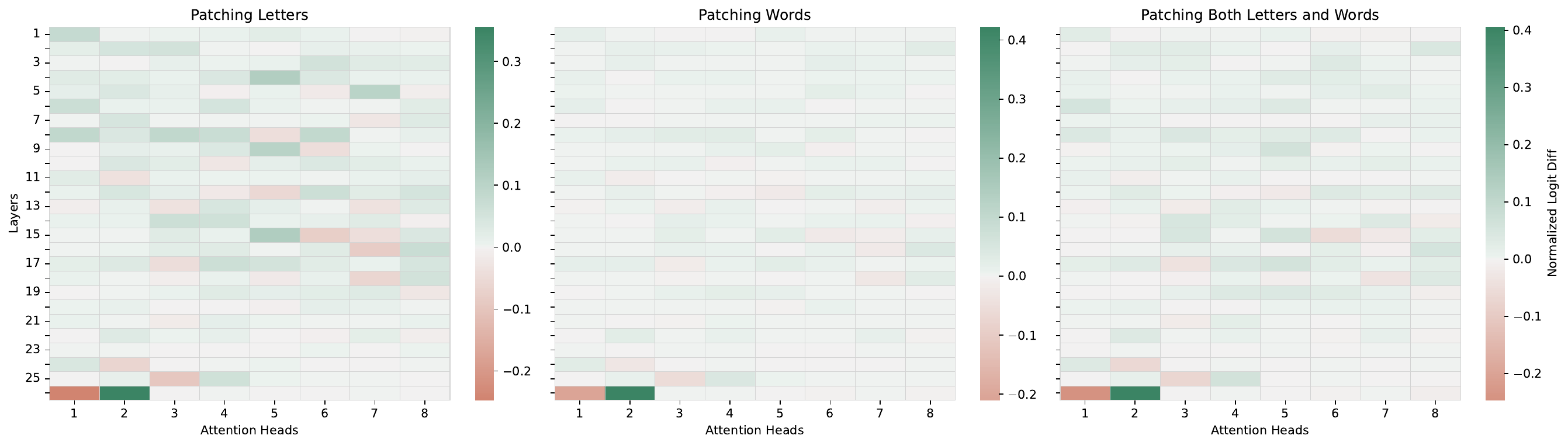}
        \caption{Gemma2-2B}
        \label{fig:gemma2b-actpatch-attention}
    \end{subfigure}
    \begin{subfigure}[d]{\textwidth}
        \centering
        \includegraphics[width=\textwidth]{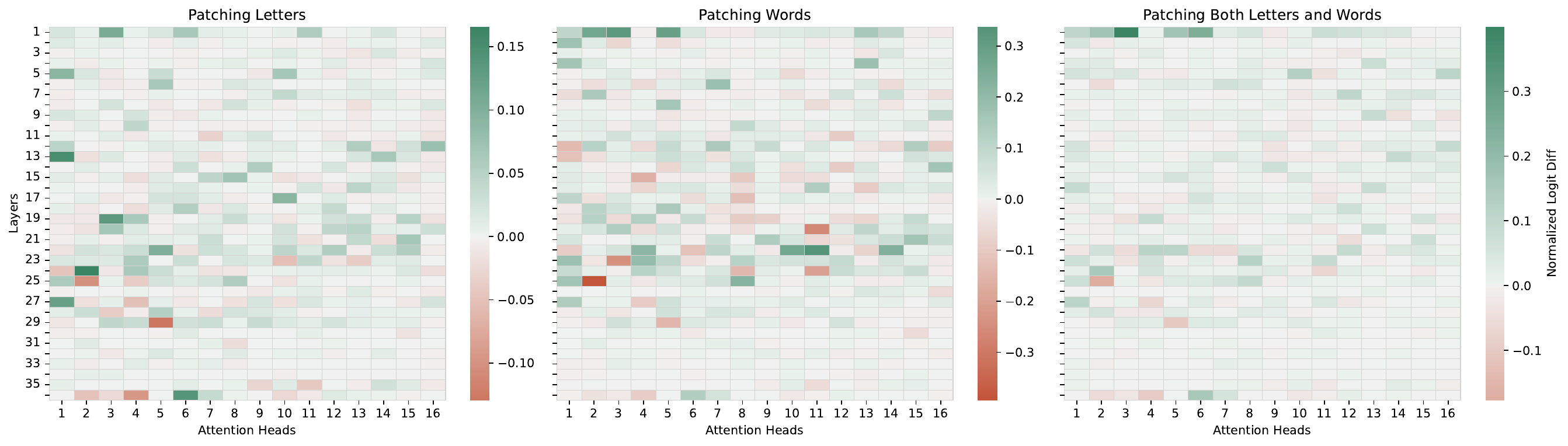}
        \caption{Gemma2-2B-Instruct}
        \label{fig:gemma2b-it-actpatch-attention}
    \end{subfigure}
    \caption{Activation patching heatmaps when corrupting the word, letter or both letter and word for Gemma attention heads averaged over 100 samples. Green regions indicate components where clean activations restore the correct behavior, helping identify localized circuits for character counting. The Red regions indicate components which diminish performance when patched in.}
    \label{fig:gemma-actpatch-attention}
\end{figure*}

We perform activation patching as described in Section~\ref{sec:act-patch-method}. Figure~\ref{fig:gemma-actpatch-combined} shows restoration heatmaps for word-level, letter-level, and  both word and letter level corruptions across residual streams, attention heads, and MLP blocks for Gemma models, and Figure~\ref{fig:gemma-actpatch-attention} shows the restoration effects on a per-attention head basis.

\section{Experiments with Qwen-2.5}
\label{sec:appendix-qwen}
To examine the generality of our findings across model scales, we replicate key experiments on the 3B and 7B parameter models (Qwen2.5-3B, Qwen2.5-7B) along with their instruction-tuned counterparts (Qwen2.5-3B-Instruct, Qwen2.5-7B-Instruct).

\subsection{Activation Patching}
\label{sec:appendix-qwen-actpatch}
\begin{figure*}[t]
    \centering
    \begin{subfigure}[a]{\textwidth}
        \centering
        \includegraphics[width=\textwidth]{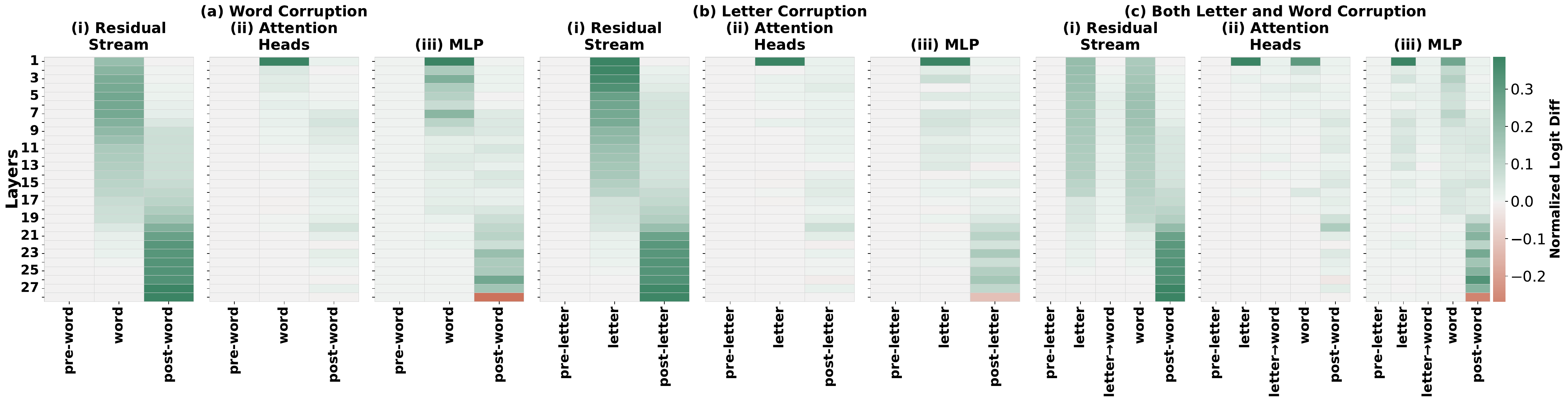}
        \caption{Qwen2.5-7B}
        \label{fig:qwen7b-act-patch}
    \end{subfigure}
    \hfill
    \begin{subfigure}[b]{\textwidth}
        \centering
        \includegraphics[width=\textwidth]{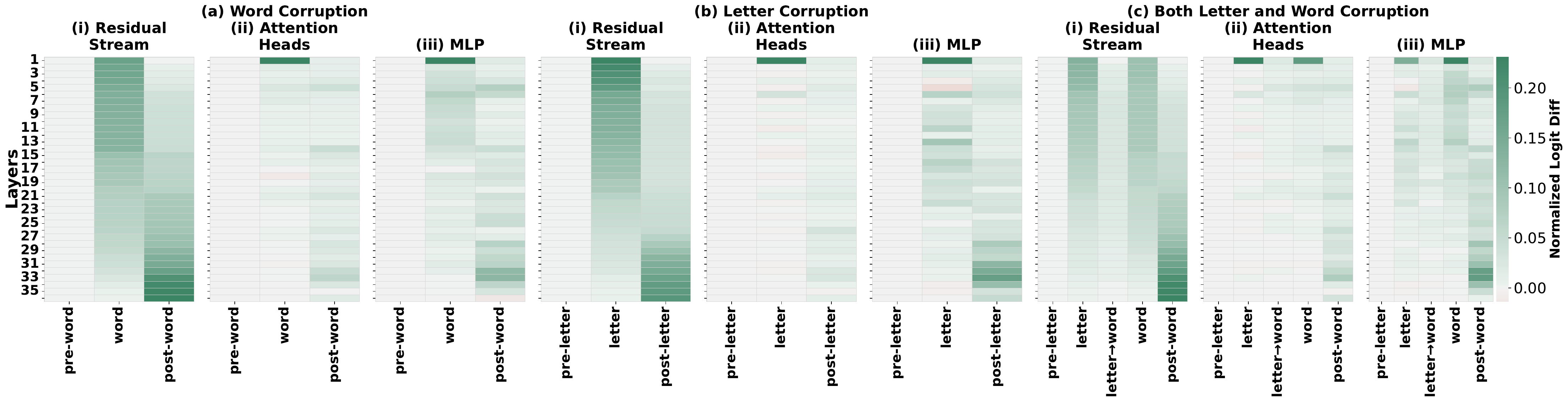}
        \caption{Qwen2.5-3B}
        \label{fig:qwen3b-actpatch}
    \end{subfigure}
    \begin{subfigure}[c]{\textwidth}
        \centering
        \includegraphics[width=\textwidth]{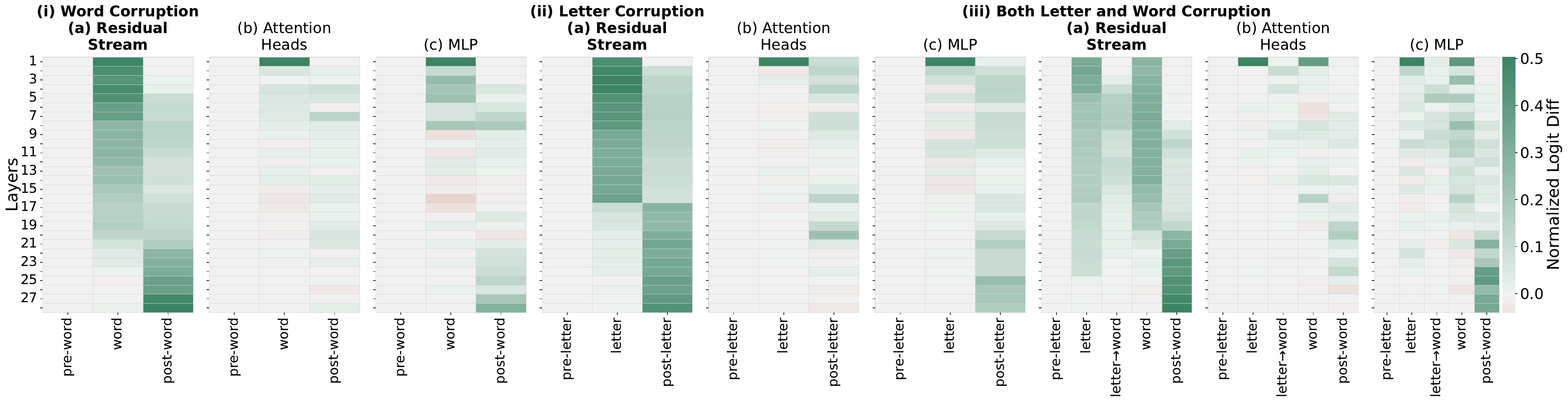}
        \caption{Qwen2.5-7B-Instruct}
        \label{fig:qwen7b-it-actpatch}
    \end{subfigure}
    \begin{subfigure}[d]{\textwidth}
        \centering
        \includegraphics[width=\textwidth]{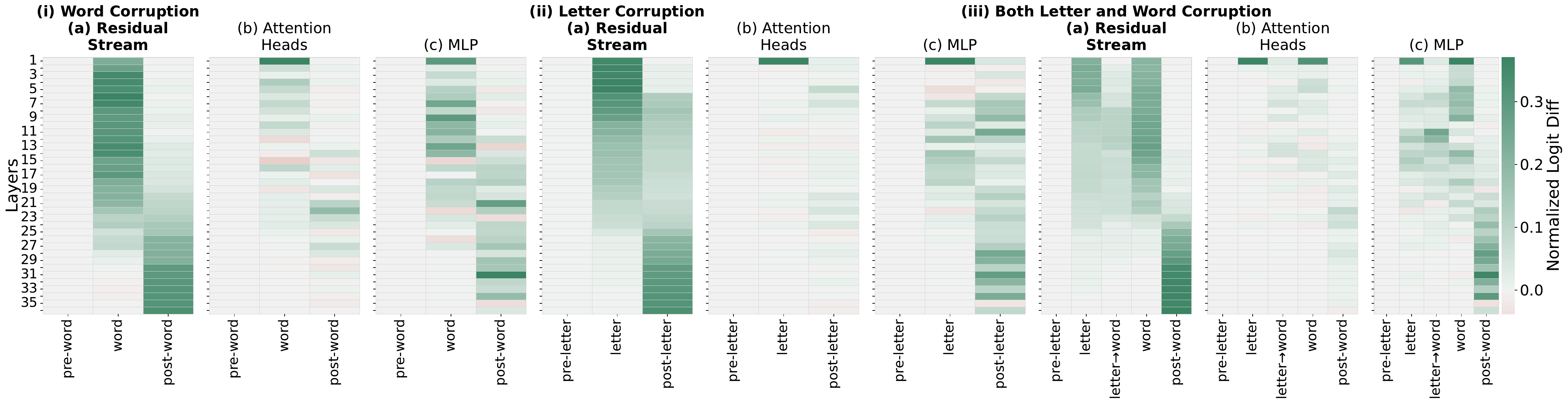}
        \caption{Qwen2.5-3B-Instruct}
        \label{fig:qwen3b-it-actpatch}
    \end{subfigure}
    \caption{Activation patching heatmaps for Qwen Models averaged over 100 samples. Green regions indicate components where clean activations restore the correct behavior, helping identify localized circuits for character counting.}
    \label{fig:qwen-actpatch-combined}
\end{figure*}

\begin{figure*}[t]
    \centering
    \begin{subfigure}[a]{\textwidth}
        \centering
        \includegraphics[width=\textwidth]{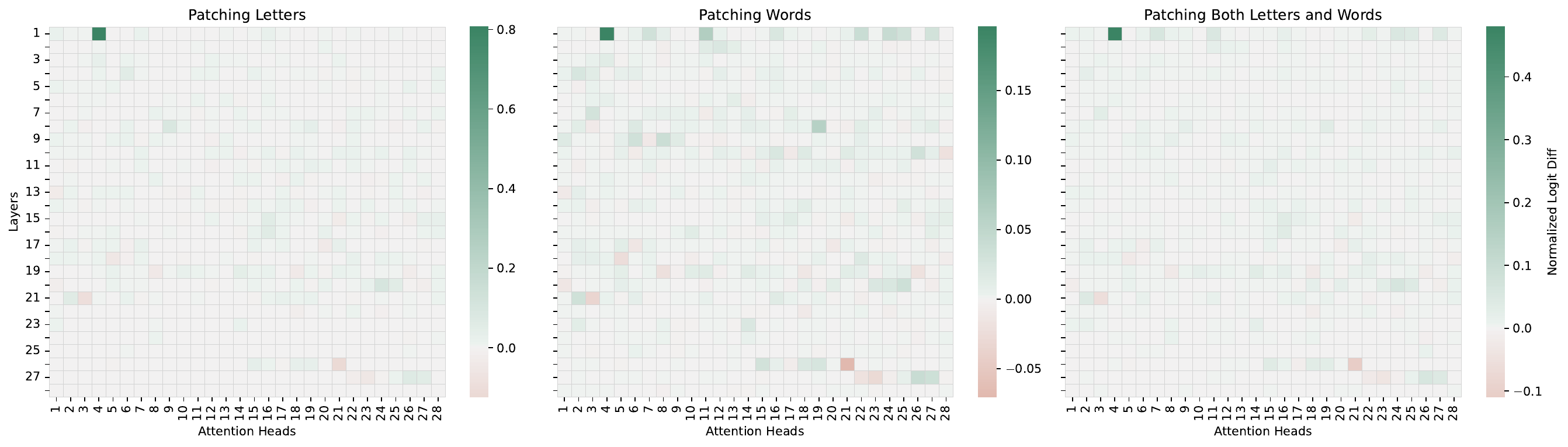}
        \caption{Qwen2.5-7B}
        \label{fig:qwen7b-act-patch-attention}
    \end{subfigure}
    \hfill
    \begin{subfigure}[b]{\textwidth}
        \centering
        \includegraphics[width=\textwidth]{figures/Qwen2.5-3B/attention_head_patch.pdf}
        \caption{Qwen2.5-3B}
        \label{fig:qwen3b-actpatch-attention}
    \end{subfigure}
    \begin{subfigure}[c]{\textwidth}
        \centering
        \includegraphics[width=\textwidth]{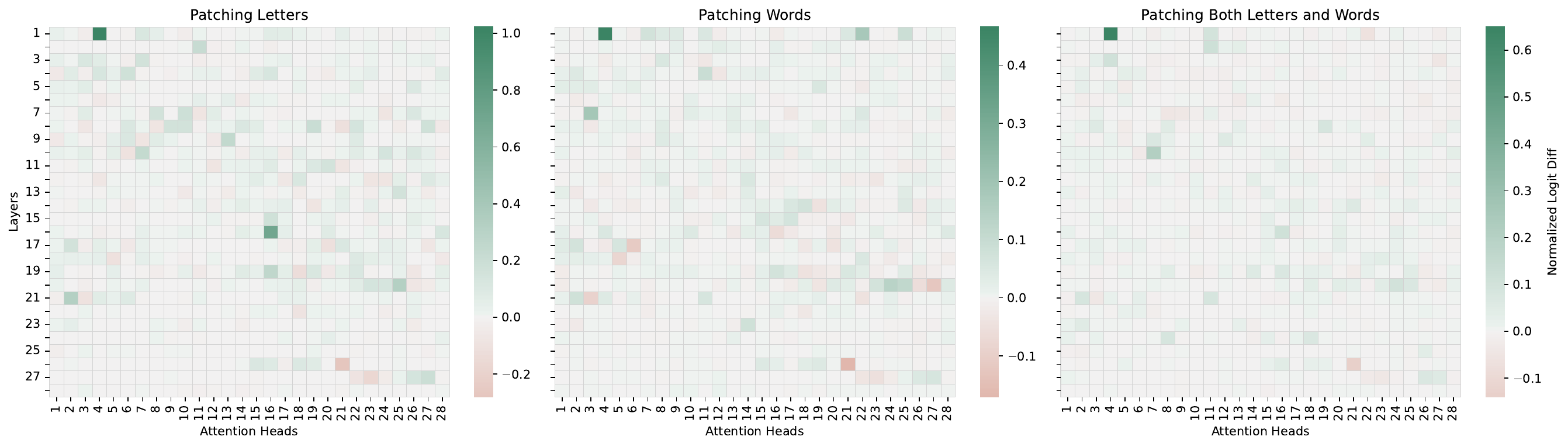}
        \caption{Qwen2.5-7B-Instruct}
        \label{fig:qwen7b-it-actpatch-attention}
    \end{subfigure}
    \begin{subfigure}[d]{\textwidth}
        \centering
        \includegraphics[width=\textwidth]{figures/Qwen2.5-3B-Instruct/attention_head_patch.pdf}
        \caption{Qwen2.5-3B-Instruct}
        \label{fig:qwen3b-it-actpatch-attention}
    \end{subfigure}
    \caption{Activation patching heatmaps when corrupting the word, letter or both letter and word for Llama attention heads averaged over 100 samples. Green regions indicate components where clean activations restore the correct behavior, helping identify localized circuits for character counting. The Red regions indicate components which diminish performance when patched in.}
    \label{fig:qwen-actpatch-attention}
\end{figure*}

We perform activation patching as described in Section~\ref{sec:act-patch-method}. 
Figure~\ref{fig:qwen-actpatch-combined}, shows restoration heatmaps for word-level, letter-level, and  both word and letter level corruptions across residual streams, attention heads, and MLP blocks for Qwen2.5 Models and Figure~\ref{fig:qwen-actpatch-attention} shows the restoration effects on a per-attention head basis.  

\end{document}